\documentclass[final,5p,times,twocolumn]{elsarticle}

\usepackage{amssymb}
\usepackage{graphicx} 

\usepackage{array}

\usepackage{latexsym}
\usepackage{multirow}
\usepackage{amsmath}
\usepackage{comment}
\usepackage{tabularx}
\usepackage{booktabs}
\usepackage{float}
\usepackage{subcaption}
\usepackage{pdflscape}
\usepackage{graphicx}
\usepackage{longtable}
\usepackage{adjustbox}
\usepackage[table]{xcolor}
\usepackage{lscape}
\usepackage{pdfpages}
\usepackage{afterpage}
\usepackage{capt-of}
\usepackage{graphicx}

\usepackage{booktabs}  

\journal{arXiv}

\begin{document}

\begin{frontmatter}


 \author{Sheresh Zahoor\corref{cor1}\fnref{label1}}
 \ead{sheresh.zahoor@mycit.ie}
 
 \cortext[cor1]{Corresponding author}
\affiliation[label1]{organization={Munster Technological University},
             addressline={Rossa Ave, Bishopstown},
            city={Cork},
            postcode={T12 P928},
            country={Ireland}}

\title{Investigating the validity of structure learning algorithms in identifying risk factors for intervention in patients with diabetes}

 

\author[label2]{Anthony C. Constantinou}

\affiliation[label2]{organization={Bayesian Artificial Intelligence Research Lab, Machine Intelligence and Decision Systems (MInDS) Group, Queen Mary University of London (QMUL)},
            city={London},
            postcode={E1
4NS}, 
            country={United Kingdom}}

\author[label3]{Tim M Curtis}
\affiliation[label3]{organization={Queen’s University Belfast},
            city={Belfast},
            country={United Kingdom}}

\author[label3,label1]{Mohammed Hasanuzzaman}

\begin{abstract}
Diabetes, a pervasive and enduring health challenge, imposes significant global implications on health, financial healthcare systems, and societal well-being. This study undertakes a comprehensive exploration of various structural learning algorithms to discern causal pathways amongst potential risk factors influencing diabetes progression. The methodology involves the application of these algorithms to relevant diabetes data, followed by the conversion of their output graphs into Causal Bayesian Networks (CBNs), enabling predictive analysis and the evaluation of discrepancies in the effect of hypothetical interventions within our context-specific case study.

This study highlights the substantial impact of algorithm selection on intervention outcomes. To consolidate insights from diverse algorithms, we employ a model-averaging technique that helps us obtain a unique causal model for diabetes derived from a varied set of structural learning algorithms. We also investigate how each of those individual graphs, as well as the average graph, compare to the structures elicited by a domain expert who categorised graph edges into high confidence, moderate, and low confidence types, leading into three individual graphs corresponding to the three levels of confidence.

The resulting causal model and data are made available online, and serve as a valuable resource and a guide for informed decision-making by healthcare practitioners, offering a comprehensive understanding of the interactions between relevant risk factors and the effect of hypothetical interventions. Therefore, this research not only contributes to the academic discussion on diabetes, but also provides practical guidance for healthcare professionals in developing efficient intervention and risk management strategies.

\end{abstract}

\begin{keyword}

Causal discovery; Causal Bayesian Networks; Non-Communicable Diseases; Diabetes

\end{keyword}

\end{frontmatter}


\section{Introduction}
\label{introduction}
Non-communicable diseases (NCDs), commonly known as chronic diseases, present a significant challenge to public health globally, impacting individuals of all ages and geographical regions. These persistent health conditions result from a complex interplay of genetic, environmental, behavioural, and physiological factors. According to data from the World Health Organization (WHO) \cite{who-ncd-mortality}, NCDs are responsible for the majority of global mortality, accounting for an estimated 41 million deaths annually, with cardiovascular diseases (CVD) leading the statistics followed by cancer, chronic respiratory diseases, and diabetes mellitus. Effective prevention and management strategies are crucial to address this growing challenge. 

The emergence of Artificial Intelligence (AI) has revealed new possibilities for transforming healthcare. AI offers immense potential in various aspects, including diagnosis, disease prediction, and treatment optimisation \cite{review1}. However, many AI systems lack transparency, hindering their interpretability and limiting their ability to capture causal relationships. To address this issue, there is a growing interest in developing models that not only provide accurate predictions but also offer transparency through causal inference, facilitating optimal decision-making under uncertainty through the simulation of hypothetical interventions. Causal inference plays a crucial role in addressing this critical gap in AI-powered healthcare solutions, where the ability to intervene effectively can significantly impact patient outcomes \cite{doi:10.1126/science.aac6076}, \cite{reason:Pearl09a}, \cite{janzing2017detecting}. It enables us to move beyond mere prediction and towards understanding the underlying cause-and-effect relationships.

Graphical models have gained popularity as a means of capturing causal relationships probabilistically \cite{Pearl1988}, \cite{spirtes2000causation}, \cite{HeckermanS95}, \cite{Dawid}. The Causal Bayesian Network (CBN) framework, introduced by Pearl \cite{10.5555/2876686.2876719}, is a widely used approach for estimating how the probabilities of a given output may change due to external interventions. Therefore, CBNs provide a powerful framework for modelling and reasoning about causal relationships in complex systems, including those found in healthcare. The graphical structure of a CBN model is represented by a Directed Acyclic Graph (DAG) $G$, which is said to be a CBN if the arcs are assumed to represent causal relationships. The literature has extensively studied learning the structure of causal networks from observational data \cite{pearl1991theory}, \cite{inbook}, \cite{spirtes1991algorithm}. The well-established classes of learning are:
\begin{itemize}
    \item \textbf{Constraint-Based Learning:} A class of learning that typically starts with a fully connected undirected graph, and employs conditional independence tests on the data to eliminate most of those edges between variables and orientate some of the remaining edges.
\item \textbf{Score-Based Learning:} This class of learning is made up of two parts: a search strategy that determines how to explore the search-space of possible graphs, and an objective function that is used to evaluate each graph visited within the search-space of graphs, affecting the paths visited by the search strategy.
\item \textbf{Hybrid algorithms:} combine the above two classes of structure learning. A typical process involves applying conditional independence tests to the data to eliminate spurious relationships, and then score-based learning to the reduced search-space of graphs.
\end{itemize}

This study investigates causal networks underlying diabetes. By gaining a better understanding of its root causes, we aim to improve diabetes management. This pursuit extends beyond mere prediction, leveraging the power of CBNs to unlock deeper insights into the factors contributing to the disease. This is because
CBNs allow us to simulate real-world interventions through the mathematical framework called \emph{do-operator} introduced by Pearl and thus allow us to answer crucial questions beyond prediction, such as, \emph{"How does education of diabetes affect the management of diabetes?"} or \emph{"What effect does blood pressure have on diabetes?"}. This capability transcends traditional observational data analysis, enabling us to truly understand cause and effect, not just correlations.

The key contributions of this research are:
\begin{itemize}

\item Developed a set of knowledge graphs for diabetes with different levels of confidence, through a collaboration with a clinical expert, that encompass important entities, relationships, and attributes that are pertinent to the disease and its associated factors. This knowledge graph provides a solid foundation for further analysis and dissemination of information in the field of diabetes research.
\item Identified causal pathways amongst potential risk factors affecting the progression of diabetes by means of an extensive analysis utilising various structural learning algorithms (constraint-based, score-based, and hybrid approaches). This method enhances the credibility and applicability of the findings compared to relying on a single algorithm, potentially leading to new insights into the fundamental causal mechanisms of diabetes.
\item Investigated the stability and consistency of identified causal pathways across different learning algorithms. This analysis provides deeper understanding into the dependability and general applicability of the results, offering better comprehension of the causal elements involved in diabetes.
\item Developed an integrated causal model for diabetes by merging insights from diverse learning algorithms by means of a model-averaging technique. This model combines the strengths of individual approaches and reduces potential biases inherent in any single algorithm, resulting in a more robust outcome. 
\item Validated and refined the inferred causal relationships by comparing the structures obtained from individual algorithms and the averaged graph with those identified by domain experts. This integrates domain knowledge into the research, enhancing the credibility and potential impact of the findings. Furthermore, by exploring areas of agreement and divergence between expert-driven models and algorithm-derived structures, the research identifies potential areas for improvement and refinement in the future iterations of the causal model.
\item Demonstrated the practical application of the research findings by utilising CBNs to evaluate the potential effects of hypothetical interventions on diabetes outcomes. This analysis delivers valuable, evidence-based information to inform public health decision-making regarding diabetes prevention and management strategies. By examining the effectiveness of different intervention policies and their implications, the research offers insights to support the development and implementation of more effective strategies for combating diabetes.
\end{itemize}

The paper is structured as follows: we discuss relevant works in Section \ref{related}, we describe the dataset and provide exploratory analysis in Section \ref{data and exp}, we describe the methodology in Section \ref{method}, we present and discuss the results in Section \ref{results}, and Section \ref{conclusion} provides our conclusions.

\section{Related works}
\label{related}
This section provides an overview of relevant literature on ML and BNs in healthcare. By exploring works in both fields, it is possible to gain a comprehensive understanding of their respective strengths and limitations.
Research on disease prediction has advanced significantly, covering various aspects such as diagnosis, prediction, categorisation, and treatment. 

ML and AI techniques have been widely investigated and applied in various healthcare domains. These include genomics \cite{genome}, treatment selection \cite{treatment}, outcome, prognosis, and prediction \cite{prognosis}. The primary objective of disease diagnosis is to identify patterns in a given dataset, and the use of ML in healthcare diagnostics and clinical decision-making is becoming increasingly popular. In their research on the early-stage risk prediction of NCDs, as presented in \cite{9469813}, the authors introduce a novel ML-based health Cyber-Physical System (CPS) framework to tackle the challenges associated with effectively processing wearable IoT sensor data for early risk prediction of diseases like diabetes. Although their work primarily focuses on the early-stage risk prediction using ML algorithms, the results do not take into account certain factors that could significantly impact disease progression. 

In \cite{roy2023prevalence}, the prevalence of NCDs across gender and age groups was examined using descriptive statistical methods. Additionally, the study explored the cumulative effects of various combinations of NCDs. Moreover, \cite{diwedi} conducted a review of six ML algorithms to predict heart disease, with logistic regression producing the highest classification accuracy. The study focused mainly on disease prediction and evaluating the effectiveness of different algorithms.
In \cite{PrakashZHDLQLF16}, memory networks were used to diagnose inference using free text clinical records and an external knowledge source, while in \cite{Shawang} a hierarchical GRU-based neural network was proposed to predict clinical outcomes based on medical code sequences from previous patient visits. These studies address different challenges in healthcare using ML and AI, ranging from sequential disease prediction to cardiovascular disease diagnosis using covariates such as sex, blood pressure, and cholesterol levels.

BNs have attracted growing research attention in healthcare and many BN models have been developed for medical diagnosis, incorporating data, knowledge, medical literature, ontologies, and other sources of medical evidence, either individually or in combination. Relevant applications of BNs include diagnosing breast cancer \cite{breastcancer}, age-related cognitive degeneration diagnoses \cite{agerelated}, and understanding severity of a patient’s condition \cite{patientsevere}. In \cite{KITSON2021103588}, publicly available Demographic and Health Survey (DHS) data is combined with knowledge to construct CBNs and investigate factors associated with infantile diarrhoea. A similar approach is taken in \cite{irina}, where a CBN model is employed to analyse a diabetic case study. The model is built and evaluated using the Netica software based on three main factors, providing insights into the emergence of diabetes and associated major problems. \cite{cbnkat} describes a strategy for integrating various sources of empirical data and verified theoretical models to improve both qualitative and quantitative HRA applications using the BN model. The use of causal independence is demonstrated in \cite{541341} as a way to simplify probability assessment and inference in BNs. \cite{cao} also highlight that BNs are useful for predicting long-term health-related quality of life and comorbidities in bariatric surgery patients. Lastly, in \cite{Spyroglou}, it was demonstrated that using BN classifiers for predicting asthma exacerbations has an advantage over traditional clinical prediction methods due to the use of multiple factors associated with exacerbations simultaneously.

 \section{Data and exploratory analysis}
\label{data and exp}
Data was obtained and pre-processed from the Behavioral Risk Factor Surveillance System (BRFSS), which is the primary system of health-related telephone surveys that collect state data on risk behaviours, chronic health conditions, and use of preventative treatments amongst U.S. residents \cite{centers}. The survey started in 1984 and currently performs over 400,000 adult interviews each year, making it the world's largest continuously conducted health survey system. This survey data provides a dataset that could be used to analyse and forecast diabetes risk variables. We utilised the BRFSS-2015 dataset, which included 253,680 health assessments. 

\subsubsection{Data pre-processing}
\label{data preprocessing}
We pre-processed the data and selected 22 variables that are relevant to this study. We present these variables in Table \ref{tab:variables}and group them as follows:
\begin{itemize}
\item \textbf{Non-modifiable RFs (Risk Factors)}: Sex, Age.
\item \textbf{Modifiable RFs:} Body mass index (BMI), Blood Pressure (HighBP), Cholesterol (HighChol), General Health (GenHealth), Education, Heart Disease (HeartDiseaseorAttack), Physical Activity (PhysActivity).
\item \textbf{Medical condition:} Diabetes (Diabetes\_binary).
\end{itemize}

\begin{table}
\small
    \centering
    \caption{The 22 variables collated, their values, and frequencies of those values.}
    \label{tab:variables}
    \begin{tabular}{{p{3cm}p{2.5cm}p{2.5cm}}}
        \hline
        Variable & States & Marginal \\
        \hline
        Diabetes\_binary & \{0 No, 1 Yes\} & \{86\%, 14\%\} \\[1ex]
        HighBP & \{0 No, 1 Yes\} & \{57\%, 43\%\} \\[1ex]
        HighChol & \{0 No, 1 Yes\} & \{58\%, 42\%\} \\[1ex]
        BMI & \{0 0-24, 1 25-39, 2 $\geq$ 40\} & \{28\%, 66\%, 6\%\} \\[1ex]
        HeartDiseaseOrAttack & \{0 No, 1 Yes\} & \{91\%, 9\%\} \\[1ex]
        CholCheck & \{0 No, 1 Yes\} & \{4\%, 96\%\} \\[1ex]
        Stroke & \{0 No, 1 Yes\} & \{96\%, 4\%\} \\[1ex]
        Smoker & \{0 No, 1 Yes\} & \{56\%, 44\%\} \\[1ex]
        Fruits & \{0 No, 1 Yes\} & \{37\%, 63\%\} \\[1ex]
        Veggies & \{0 No, 1 Yes\} & \{19\%, 81\%\} \\[1ex]
        HvyAlcoholConsump & \{0 No, 1 Yes\} & \{94\%, 6\%\} \\[1ex]
        AnyHealthcare & \{0 No, 1 Yes\} & \{5\%, 95\%\} \\[1ex]
        NoDocbcCost & \{0 No, 1 Yes\} & \{92\%, 8\%\} \\[1ex]
        MentHlth & \{0, 1, 2\} & \{87\%, 5\%\, 7\%\} \\[1ex]
        PhysHlth & \{0, 1, 2\} & \{84\%, 5\%\, 10\%\} \\[1ex]
        DiffWalk & \{0 No, 1 Yes\} & \{83\%, 16\%\} \\[1ex]
        Sex & \{0 Female, 1 Male\} & \{56\%, 44\%\} \\[1ex]
        Age & \{1, 2, 3, 4, 5, 6\} & \{5\%, 10\%, 14\%, 23\%, 26\%, 22\%\} \\[1ex]
        Income & \{1, 2, 3, 4\} & \{8\%, 14\%, 25\%, 53\%\} \\[1ex]
        Education & \{1, 2, 3\} & \{2\%, 29\%, 70\%\} \\[1ex]
        GenHlth & \{1, 2, 3\} & \{53\%, 42\%, 5\%\} \\[1ex]
        PhysActivity & \{0 No, 1 Yes\} & \{25\%, 75\%\} \\[1ex]
        \hline
    \end{tabular}
\end{table}

In the analysis of the BRFSS data, certain pre-processing steps were essential to ensure compatibility with various algorithms that require categorical data for constructing CBNs. This involved encoding non-categorical variables, since some variables within the dataset were not categorical in nature. It also involved ensuring that existing categorical variables are restricted to a reasonable number of states, since many of the variables had a large number of possible states which can be impractical for constructing CBNs. Therefore, the next step involved reducing the number of states while ensuring that no essential information was lost for the purposes of this case study. Specifically, :
\begin{enumerate}
\item\textbf{Age grouping:} Age was categorised into six groups: 18-29, 30-44, 45-54, 55-64, 65-74, and 75 or older. These groups were further encoded numerically as 1, 2, and so on.

\item\textbf{General health:} General health, which initially had several states, was simplified into three categories: excellent (1), good (2), and poor (3).

\item\textbf{Education:} Education levels were condensed into three states: individuals who never attended school or attended from grade 1 through grade 8 (1), those who completed grade 9 through GED (2), and individuals with college education (3).

\item\textbf{Income:} Income levels were condensed into four states: individuals with income between \$0-14,999 as 1, between \$15,000-24,999 as 2, between \$25,000-50,000 as 3 and greater than \$50,000 as 4.

\item\textbf{BMI categorisation:} BMI, initially continuous, was categorised into three states: 0-24 (0), 25-39 (1), and $\geq 40$ (2).

\item\textbf{Physical and mental health duration:} Variables related to physical and mental health, which had values ranging from 0 to 30, were transformed into three categories: 0-9 days (0), 10-19 days (1) and $\geq 20$ (3). 
\end{enumerate}

\subsubsection{Exploratory Data Analysis}
\label{exploratory}
\begin{figure}[H]
\begin{center}
    \includegraphics[width=8cm]{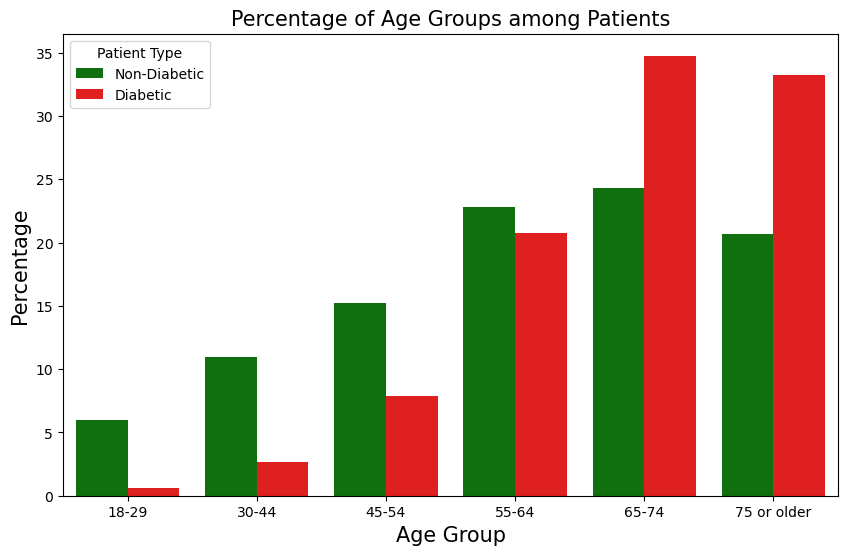}
    \caption{Age distribution of diabetic and non-diabetic individuals.}
    \label{fig:age}
   \end{center}
\end{figure}

\begin{figure}[H]
\begin{center}
    \includegraphics[width=8cm]{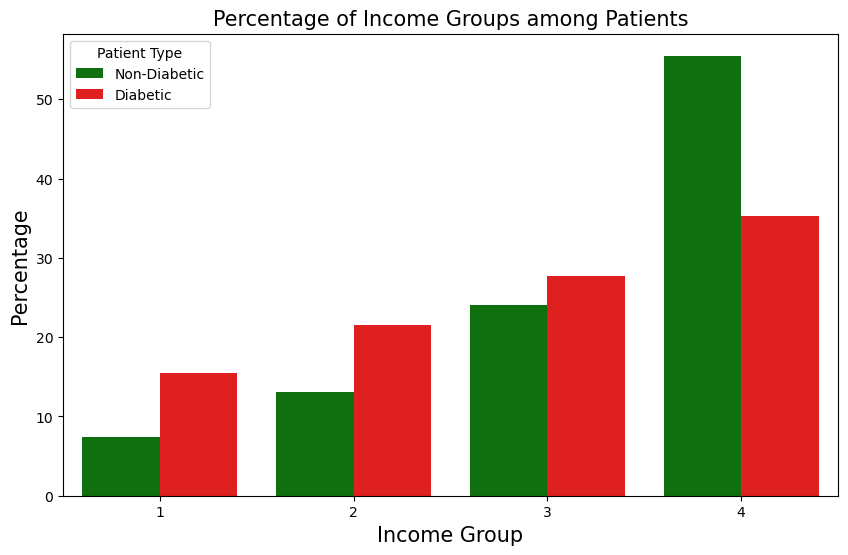}
    \caption{Annual income distribution of diabetic and non-diabetic individuals.}
    \label{fig:income}
     \end{center}
\end{figure}

\begin{figure}[H]
\begin{center}

    \includegraphics[width=8cm]{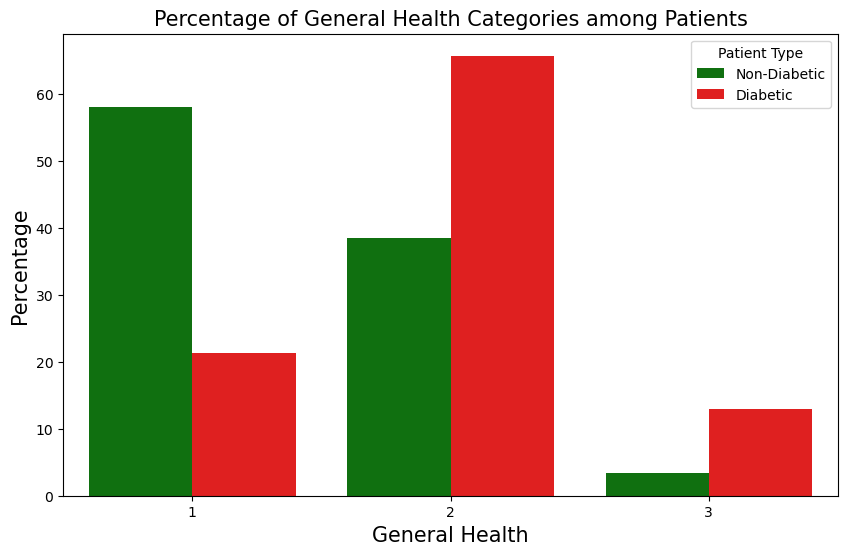}
    \caption{General Health distribution of diabetic and non-diabetic individuals.}
    \label{fig:generalhealth}
    \end{center}
\end{figure}

\begin{figure}[H]
\begin{center}

    \includegraphics[width=8cm]{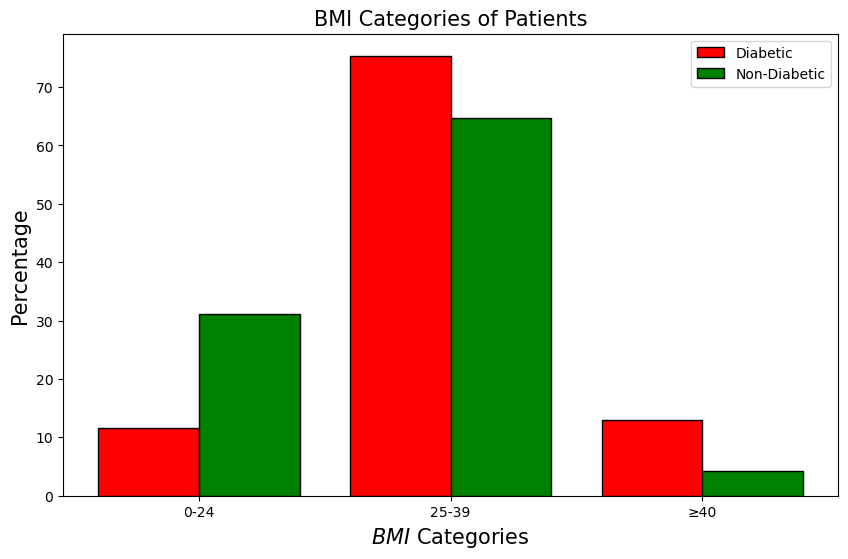}
    \caption{BMI distribution of diabetic and non-diabetic individuals.}
    \label{fig:bmi}
    \end{center}
\end{figure}

\begin{figure}[H]
\begin{center}
    \includegraphics[width=8cm]{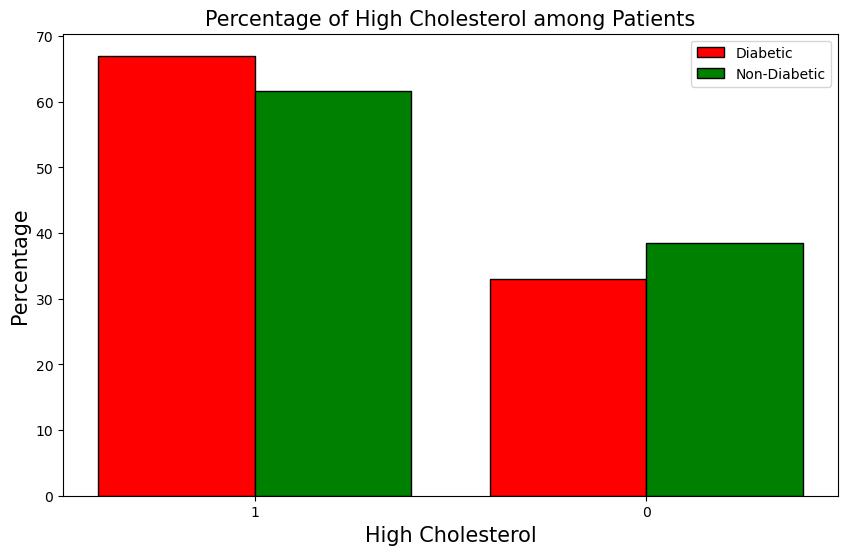}
    \caption{High Cholesterol distribution of diabetic and non-diabetic individuals.}
    \label{fig:cholesterol}
    \end{center}
\end{figure}

\begin{figure}[H]
\begin{center}
    \includegraphics[width=8cm]{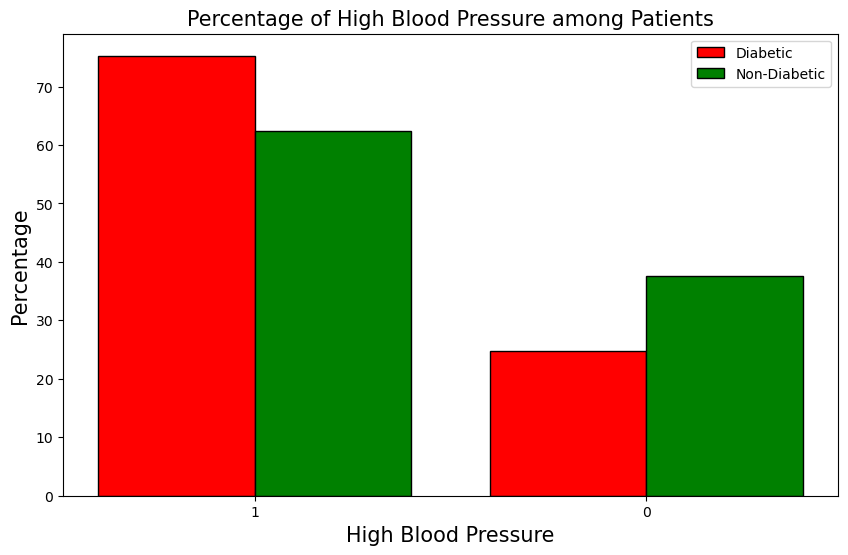}
    \caption{High Blood Pressure distribution of diabetic and non-diabetic individuals}
    \label{fig:highbp}
     \end{center}
\end{figure}

This section presents an initial exploration of the relationships between key variables and the target variable, diabetes, within our pre-processed dataset. Key observations include:
\begin{enumerate}
\item\textbf{} The age distribution (Figure \ref{fig:age}) showed a higher prevalence of diabetes in individuals aged 55 and above, aligning with existing research.
\item\textbf{} Regarding socioeconomic factors, the income levels displayed a difference between diabetic and non-diabetic groups, with a higher proportion of non-diabetics falling into higher income categories (Figure \ref{fig:income}).
\item\textbf{} The distribution of individuals across general health categories shifted, with a decrease in the "Excellent" category and an increase in the "Poor" category amongst those with diabetes (Figure \ref{fig:generalhealth}).
\item\textbf{} The distribution of individuals across BMI categories shifted towards higher levels in the diabetic group, highlighting the importance of monitoring individuals in the 25-39 BMI range (Figure \ref{fig:bmi}).
\item\textbf{} The percentage of individuals with high cholesterol significantly increased in the diabetic group compared to non-diabetics (Figure \ref{fig:cholesterol}).
\item\textbf{} The prevalence of high blood pressure was substantially higher in the diabetic group compared to the non-diabetic group (Figure \ref{fig:highbp}).
\end{enumerate}
These findings provide preliminary insights into the potential relationships between these variables and diabetes, warranting further investigation through statistical analysis and causal inference techniques.

\section{Methodology}
\label{method}

This section is separated into four parts. We begin by detailing the process we followed to obtain a causal structure with the assistance of a domain expert in Section \ref{Knowledge graph}. Following this, we elaborate on the process employed to learn causal structures solely from data in Section \ref{causal structure}. Subsequent subsections describe the model-averaging approach utilised to derive a unified structure from multiple candidate structures (Section \ref{model-average}), and the adoption of the \textit{do}-operator framework for interventional analysis (Section \ref{interventional}).
\subsection{Constructing a causal graph from knowledge}
\label{Knowledge graph}
\begin{figure*}
  \includegraphics[width=\textwidth,height=5cm]{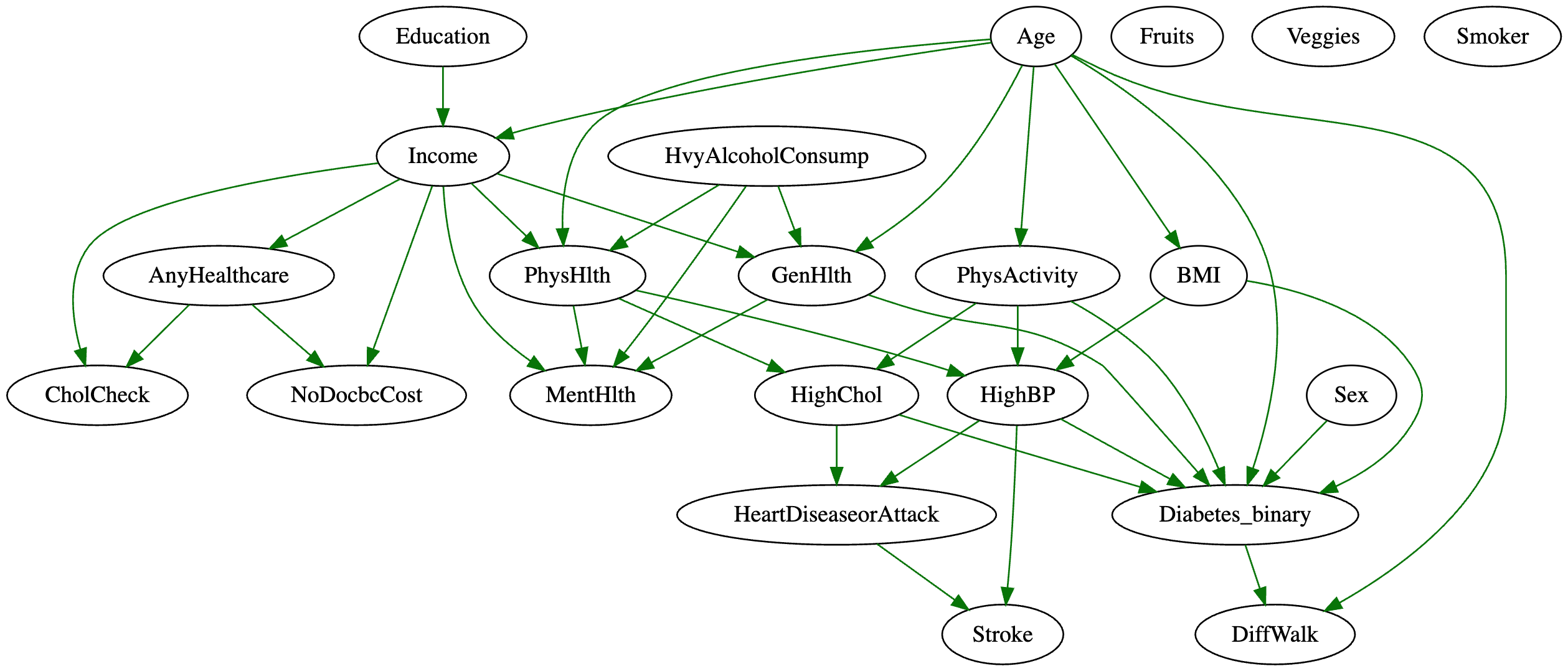}
  \caption{High confidence causal graph. It contains the 22 variables and 37 directed edges identified by the domain expert with high confidence (coloured in green).}
  \label{fig:High confidence}
\end{figure*}

\begin{figure*}

    \includegraphics[width=\textwidth,height=5cm]{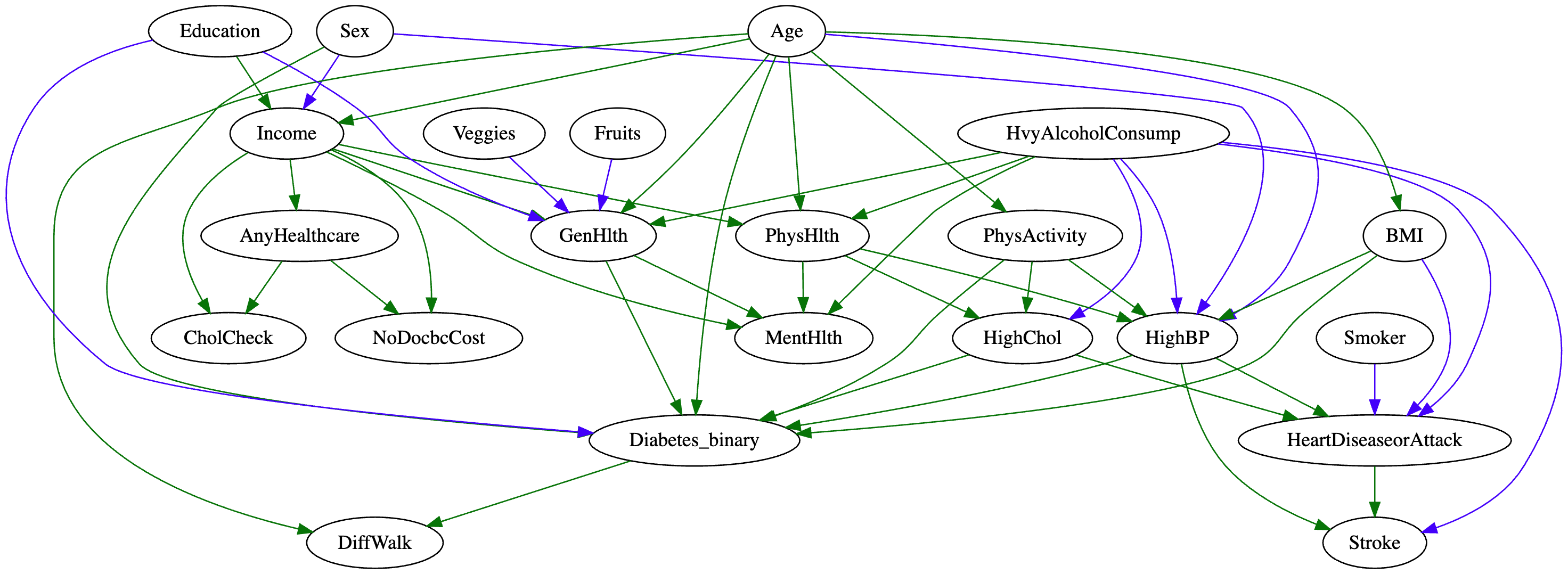}
    \caption{Moderate confidence causal graph. It contains the 22 variables and 51 directed edges identified by the domain expert with high (coloured in green) and moderate (coloured in blue) confidence.}
    \label{fig:Moderate confidence}
\end{figure*}

\begin{figure*}

    \includegraphics[width=\textwidth,height=5cm]{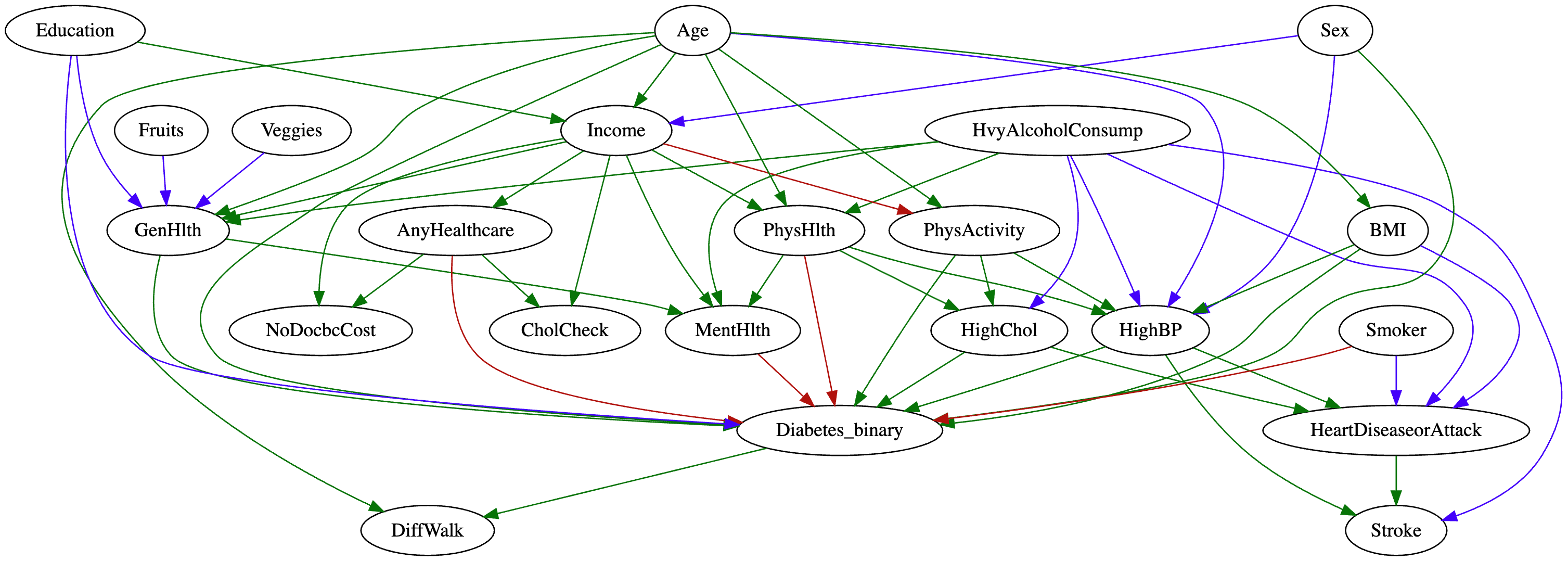}
    \caption{Low confidence causal graph. It contains the 22 variables and 56 directed edges identified by the domain expert with low confidence, where green, blue and red colours represent high, moderate and low confidence respectively.}
    \label{fig:Low confidence}
\end{figure*}

We initiated the graph construction process through collaboration with a domain expert, jointly crafting a causal graph. The domain expert identified and categorised possible causal relationships, between the variables collated, into three levels of confidence: high confidence, moderate confidence, and low confidence. We used these three levels of confidence to produce three different graphs:
\begin{enumerate}

\item\textbf{ High confidence graph:} Figure \ref{fig:High confidence} presents the causal graph containing the causal relationships with a high confidence level, represented by green-coloured edges. It provides a structured view of the most trusted connections within the knowledge domain.
\item\textbf{ Moderate confidence graph:} Figure \ref{fig:Moderate confidence} presents the causal graph containing both high and moderate confidence levels, where blue-coloured edges represent moderate confidence levels. It offers insights into relationships where confidence levels are at least moderate.
\item\textbf{ Low confidence graph:} Figure \ref{fig:Low confidence} presents the causal graph containing all edges, including those with low confidence coloured in red. It provides a comprehensive overview of potential connections, even when confidence is low.
\end{enumerate}


To strengthen the credibility of the pre-processed dataset, as well as the high-confidence graph developed by the domain expert, we employed statistical analysis to explore the strength of the potential relationships, after they had been identified by the expert. The analysis showed that the relationships share strong associations, highlighting the propagation of inference through the interconnectedness of factors within the studied population. For example, a strong correlation emerged between heavy alcohol consumption and physical health, indicating that higher alcohol intake associates with different physical health outcomes. Similarly, a significant association was identified between sex and diabetes, suggesting a potential role of gender in diabetes prevalence. Furthermore, age exhibited substantial correlations with numerous factors (income, general health, diabetes, BMI, physical activity, difficulty walking, and physical health), all statistically significant (p-values = 0.0). These preliminary analysis might emphasise the importance of considering demographic and lifestyle factors when investigating health outcomes.

While this analysis has helped us to obtain a better understanding of the associational strength between some variables of interest, it is important to clarify that correlation cannot be used to judge causal relationships. This is not only because correlations can be spurious, but also because the absence of correlation does not necessarily indicate the absence of causation.

\subsection{Learning causal structure from data}
\label{causal structure}
We employ structure learning algorithms from various learning classes to learn graphical structures. Specifically, we consider constraint-based, score-based, and hybrid learning algorithms founded on varying assumptions about the input data, and produce diverse types of graphical structures. To ensure a comprehensive exploration of the learning algorithms, our analysis encompasses a total of 14 algorithms across the three different classes of learning. This variety in algorithmic approaches aims to strengthen the robustness and generalisability of our findings.

By comparing the results obtained from these diverse methodologies, we gain a more comprehensive understanding of the underlying relationships within our data. This approach helps to mitigate potential biases that might be introduced by any single algorithm, leading to a more reliable and representative picture of the true structure. Table \ref{table:2} lists the 14 algorithms along with the package we used to execute them, and the learning class they belong to.  The 14 algorithms are:
\begin{itemize}
\item \textbf{Peter-Clark algorithm (PC):} This algorithm was proposed by Peter-Clark in 1991 \cite{spirtes1991algorithm}. It starts by forming a fully connected undirected graph and eliminates edges using conditional independence tests. It identifies the causal relationships between variables based on the absence of subsets of variables that render them independent when conditioned upon. 
\item \textbf{PC-Stable:} The PC-Stable algorithm \cite{JMLR} is an extension of the PC algorithm. It modifies PC to address the issue of its dependence on the order of variables as read from data. These modifications eliminate or reduce this dependency.
\item\textbf{Fast Causal Inference (FCI):} FCI, developed by Spirtes et al. in 1999 \cite{spirtes2000causation}, is a constraint-based method that extends the PC algorithm to account for latent variables in the data. It begins by establishing the skeleton and then uses conditional independence tests to orientate some of the edges, including the identification of bi-directed edges to indicate confounding. 
\item \textbf{Fast Greedy Equivalence Search (FGES):} FGES is an efficient parallelised version of the Greedy Equivalence Search (GES) algorithm introduced by Chickering and Meek (2002) \cite{chickering2002optimal}. Developed by Ramsey et al. (2016) \cite{fges}, FGES operates in two phases: a forward search, where edges maximising the objective function in the Completed Partially Directed Acyclic Graph (CPDAG) space are added iteratively, and a backward search, where edges are removed to enhance the objective score until no further improvement is possible.
\item\textbf{SaiyanH:} A hybrid algorithm by \cite{saiyanh} that starts with marginal and conditional independence tests to learn a sparse undirected graph, and then uses the results from those tests together with BIC and \textit{do}-operator tests to orientate the edges. The algorithm then utilises a tabu search with the restriction not to visit graphs that contain disjoing subgraphs. This learning approach guarantees that the acquired model does not contain disjoint subgraphs or variables, to enable full propagation of evidence .
\item\textbf{Hill-Climbing (HC):} HC algorithm \cite{margaritis} typically starts with an empty graph and greedily searches the space of graphs by adding, removing and reorienting edges iteratively. It uses a scoring function to score each graph visited and, at each iteration, moves to the graph that maximises the scoring function. It terminates when no neighbouring graph further increases the score.
\item\textbf{Structural Expectation Maximisation (EM):} Structural EM \cite{structem} combines the traditional EM approach for parameter optimisation with score-based learning for model selection. The Structural EM  trains networks using penalised likelihood scores, which include the BIC/MDL score and several Bayesian score estimates.
\item\textbf{Grow-Shrink (GS):} The GS algorithm \cite{Dimitri} is a method for discovering a variable's Markov Blanket, MB(X), which includes all of the variables in a graph that hold information about X; i.e., its parents, its children, and the parents of its children. The algorithm has two stages: a) adds variables to an empty set S based on their relationship with X, and b) removes variables that do not belong to MB(X). The GS algorithm then uses those MBs to construct a global graph.
\item\textbf{Incremental Association Markov Blanket (IAMB):} The IAMB algorithm \cite{IAMB} involves two phases: forward and backward. During the forward phase, a candidate set (CMB) is populated with variables, potentially including false positives. This addition is based on a heuristic function \textit{f} that maximises the association between variables and the target variable (T), aiming to estimate the MB of T. In the backward phase, false positives are identified and removed from CMB to ensure its equivalence to MB(T). The heuristic function calculates the Mutual Information between variables for soundness. This strategy prioritises efficiency in both time and sample size, excluding irrelevant variables and minimising the sample required for conditional independence tests in backward conditioning.
\item\textbf{fast-IAMB:} fast-IAMB \cite{fastIamb} is an extension of IAMB that, in the growing phase, it sorts attributes from most to least conditionally dependent, and multiple attributes are added at once to reduce computational complexity and make the algorithm faster. This is different from IAMB, which adds only one attribute at a time and sorts the remaining attributes after each modification. Fast-IAMB's approach is said to be more statistically appropriate and adds more true members to the MB.  
\item\textbf{Max-Min Hill-Climbing (MMHC):} MMHC is a hybrid algorithm that first learns an undirected graph using the Max-Min Parents and Children (MMPC) approach, which is a local discovery method described in Tsamardinos et al. (2003b)\cite{mmhc}. It then uses a greedy Bayesian-scoring hill-climbing search to orientate the edges.  
\item\textbf{H2PC:} The hybrid learning H2PC algorithm \cite{h2pc} first identifies the parents and children set for each variable and then conducts a greedy hill-climbing search in the space of graphs. 
\item\textbf{Model Averaging Hill-Climbing (MAHC):} MAHC \cite{constantinou2022effective} method extends HC in two ways. First, it pre-processes the Candidate Parent Sets (CPSs) for each variable and performs pruning on CPSs to dramatically reduce the search-space of graphs. It then performs model-averaging in the hill-climbing search process, and moves to the neighbouring graph that maximises the objective function, on average, for that neighbour and over all its valid neighbours. This approach is said to be less sensitive to data noise or data imperfections typically present in real datasets.
\item\textbf{TABU:} TABU \cite{Bouckaert1995} is an extension of HC that permits the exploration of DAGs that slightly decrease the objective score, in an effort to escape local maxima solutions, and maintains a tabu list containing the most recently visited graphs to prevent the algorithm from returning to a graph that was previously visited. This approach enables TABU to move into new graphical regions that may contain an improved local maximum compared to HC. 
\end{itemize}
As shown in Table \ref{table:2}, we made use of three structure learning tools, the TETRAD \cite{ramsey2018tetrad}, Bayesys \cite{anthony}, and bnlearn \cite{bnlearn}, to apply the 14 algorithms described above to our data and obtain graphical structures. Additionally, we utilised Bayesys for model-averaging (described in the next subsection) and converting graphical structures into BN models, which were subsequently analysed using the GeNIe BN software \cite{genie}.  

However, structure learning from observational data often leads to a graphical structure known as CPDAG, which contains both directed and undirected edges. The undirected edges in a CPDAG represent relationships for which causal directionality cannot be established from observational data alone and hence, a CPDAG represents a set of DAGs that belong to the same Markov Equivalence Class (MEC). Some constraint-based algorithms may also produce a Partially Directed Acyclic Graph (PDAG) where only v-structures are captured and hence, it contains a higher number of undirected
edges than a CPDAG, and a PDAG cannot always be converted into a CPDAG or a DAG structure.

To manage computational resources efficiently, we imposed a time-limit of four hours on the execution of structure learning. The PC and FCI algorithms, implemented in TETRAD, are the only ones that exceeded this time-limit and so we could not include these algorithms in the results. Additionally, the PC-Stable algorithm, in bnlearn, yielded a PDAG featuring conflicting v-structures and hence, we could not convert its graphical structure into a CPDAG or a DAG. The inability to transform PDAGs into acyclic graphs rendered results derived from PC-Stable unsuitable for evaluation. Therefore, we obtained suitably learnt outputs from 11 out of the 14 algorithms investigated. Moreover, not all the algorithms produce a DAG, which is a necessary requirement for converting a graph into a CBN and performing causal inference. CPDAG outputs were converted into a random DAG from their corresponding Markov equivalence class.

\begin{table}
\small
    \centering
    \caption{The structure learning algorithms considered in this study.}
    \label{table:2}
    \begin{tabular}{llll}

        \hline
        Algorithm     & Package & Reference                        & Learning Class   \\
        \hline
        \\
        PC            & TETRAD  & \cite{spirtes1991algorithm}      & Constraint-based \\
        PC-Stable     & bnlearn & \cite{JMLR}                      & Constraint-based \\
        FCI           & TETRAD  & \cite{spirtes2000causation}      & Constraint-based \\
        FGES          & TETRAD  & \cite{fges}                      & Score-based      \\
        SaiyanH       & Bayesys & \cite{saiyanh}                   & Hybrid           \\
        HC            & bnlearn & \cite{margaritis}                & Score-based      \\
        Structural EM & bnlearn & \cite{structem}                  & Score-based      \\
        GS            & bnlearn & \cite{Dimitri}                   & Constraint-based \\
        IAMB          & bnlearn & \cite{IAMB}                      & Constraint-based \\
        fastIAMB      & bnlearn & \cite{fastIamb}                  & Constraint-based \\
        MMHC          & bnlearn & \cite{mmhc}                      & Hybrid           \\
        H2PC          & bnlearn & \cite{h2pc}                      & Hybrid           \\
        MAHC          & Bayesys & \cite{constantinou2022effective} & Score-based      \\
        TABU          & bnlearn & \cite{Bouckaert1995}             & Score-based \\
        \hline
    \end{tabular}
\end{table}

\subsection{Model-averaging}
\label{model-average}
Because the learnt graph could be highly sensitive to the selection of the structure learning algorithm or domain expert, we also explore the use of a model-averaging approach to consolidate insights from multiple structure learning algorithms. In the context of structure learning, model-averaging aims to obtain a comprehensive portrayal of the underlying causal structure given the data, rather than relying on a single best-performing model as identified by a single algorithm. Therefore, model-averaging addresses the inherent variations in individual algorithms by mitigating the risk of overfitting, thereby reducing variance in causal estimates and enhancing the overall reliability of the results.

We adopted the model-averaging strategy implemented in Bayesys \cite{anthony}, which takes as input multiple input graphical structures of any type, and works as follows: 
\begin{itemize}
\item[1] Begin adding directed edges to the average graph, starting with the edges that occur most frequently across input graphs.
\begin{itemize}
    \item[a)] If an edge has already been added in the reverse direction, skip it.
    \item[b)] If adding an edge would create a cycle, reverse the edge and add it to the edge-set C.
    \end{itemize}
\end{itemize}
\begin{itemize} 
\item[2]  Add undirected edges to the average graph, starting with the edges that occur most frequently.
\begin{itemize}
    \item[a)] Skip an undirected edge if it has already been added as a directed edge.
     \end{itemize}
\end{itemize}
\begin{itemize} 
\item[3] Finally, add directed edges from the edge-set C to the average graph, starting with the edges that occur most frequently.
\begin{itemize}
    \item[a)] Skip an edge if it has already been added as an undirected edge
    \end{itemize}
\end{itemize}

\subsection{Interventional analysis}
\label{interventional}
We use the GeNIe software \cite{genie} that implements Pearl’s \emph{do}-operator, to simulate interventions on variables that could be intervened on, and assess the impact of those interventions with reference to the target variable Diabetes\_binary. 
Further to what has been discussed in Introduction, the \textit{do}-operator was initially developed by Pearl \cite{pearldo} and extended by others like Valorta and Huang \cite{huang2006valtorta}, Shpitser and Pearl \cite{shpitser2006identification}, and Tian and Pearl \cite{tian2002general}. It provides a robust mathematical framework for causal inference by enabling us to simulate hypothetical interventions on a node within the network, and manipulate its distribution to a particular state. When we intervene on node $X$, for example, we essentially disconnect all the parent nodes of $X$ from affecting it; i.e., the intervened nodes is rendered independent of its causes. 

Consider a scenario with three variables: $X$, $Y$, and $Z$, where $X \rightarrow Y$, $ Z \rightarrow X $ and $Z \rightarrow Y $.  We assert that $X = x$ serves as a cause of $Y = y$. To denote this causal concept, we perform an intervention on node $X$ within the causal DAG and set it to $x$, thereby modifying the model to become \cite{Quinn}:

\begin{center}
    $P_{hypo} (Y=y,Z)_{(X=x)} = P(Y=y,Z\mid do[X=x])
 = P(Y=y\mid Z,X=x)P(do[X=x]\mid Z)P(Z)$
\end{center}

where $P_{hypo}$ is simply the probability of hypothetical intervention. Further, if we assume that $X = x^{-}$ where $x^{-}$ represents the other state, then the difference between the two probability values i.e. 
\begin{center}
$P(Y=y \mid do[X=x])$ $-$ $P(Y=y \mid do[X=x^{-}])$
\end{center}
is known as the “causal effect difference” or “average causal effect” (ACE) \cite{glymour2016causal}. If a query $Q$ such as $Q = P(y\mid do(x), z)$ is presented, the \emph{do-operator} can systematically determine its identifiability through the above algebraic procedure.

Consider four disjoint sets of nodes $X$, $Y$, $Z$, and $W$ in a causal DAG $G$. We can obtain a graph $G_{\overline X}$ by removing all arrows pointing to nodes in $X$ from $G$, and a graph $G_{\underline X}$ by removing all arrows emerging from nodes in $X$. To denote the deletion of both incoming and outgoing arrows, we use the notation $G_{\overline X \underline Z}$. The three rules presented in Table \ref{table:3} have universal applicability to all interventional distributions that conform to the structure of graph $G$ \cite{heiss2021}.

\begin{table}
\small
    \centering
  \caption{Rules for Interventional Analysis}
  \begin{tabular}{>{\raggedright}p{0.25\linewidth} p{0.6\linewidth}}
    \toprule
    \textbf{Rule} & \textbf{Description} \\
    \midrule
    Rule 1 & If $(Y {\perp\!\!\!\perp} Z \mid X, W)_{G_{\overline X}}$, 
    \\ & then $P(y \mid do(x), z, w) = P(y \mid do(x), w)$. \\
    \hline
    Rule 2 & If $(Y {\perp\!\!\!\perp} Z \mid X, W)_{G_{\overline X \underline Z}}$, \\ &then $P(y \mid do(x), do(z), w)$ \\ & $ = P(y\mid do(x), z, w)$. \\
    \hline
    Rule 3 & If $(Y {\perp\!\!\!\perp} Z \mid X, W)_{\overline X \overline {Z(W)}}$, where $Z(W)$ is the set of Z-nodes that are not 
    ancestors of any W-node in $G_{\overline X}$, then 
    $P(y \mid do(x), do(z), w) = P(y \mid do(x), w)$. \\
    \bottomrule
  \end{tabular}
  \label{table:3}
\end{table}

\subsection{Graphical evaluation}
\label{graph analysis}
 We use three metrics to compare the similarity between DAG structures obtained by the algorithms, the model-averaging process, and the domain expert. These are:
\begin{itemize}
\item[i)]\textbf{Structural Hamming Distance (SHD):} SHD is a fundamental metric used to evaluate the structural dissimilarity between two graphs. The SHD score is computed by calculating the count of structural modifications  needed to transform a learnt graph into the ground truth, with arc additions, deletions, or reversals. Minimising the SHD score aims for a closer alignment between the two graphical structures, while a higher score indicates greater dissimilarity.

\item[ii)]\textbf{F1:} The F1 score is a widely-used metric that considers the balance between precision and recall in evaluating the accuracy of predicted edges within a graphical structure. It is mathematically expressed as:

\begin{equation}
F1 = \frac{2 \cdot RP}{R + P}
\end{equation}

where R represents the recall, which is the proportion of true positive connections (correctly predicted edges) out of all actual positive connections, P represents precision, which is the ratio of true positive connections (correctly predicted edges) to the total positive connections predicted.

\item[iii)]\textbf{Balanced Scoring Function (BSF):} Unlike the SHD and F1 metrics that solely consider edge presence, the BSF metric takes into account the difficulty of discovering both the presence and absence of edges, to generate a score that is balanced, relative to the difficulty of discovering the presence of an edge. The BSF score is calculated using the following formula:

\[
BSF = 0.5 \left( \frac{TP}{a} + \frac{TN}{i} - \frac{FP}{i} - \frac{FN}{a} \right)
\]

where $TP$ is the number of true positives (correctly identified edges), $TN$ is the number of true negatives (correctly identified absent edges), $FP$ is the number of false positives (incorrectly identified edges that are not present in the ground truth), $FN$ is the number of false negatives (edges that are present in the ground truth but not identified by the model), $a$ is the number of edges in the assumed ground truth, and $i$ is the number of independencies in assumed ground truth calculated using the formula $i = \frac{|V|(|V| - 1)}{2}$, where $|V|$ is the number of variables.
\end{itemize}

The BSF score has a range from -1 to 1, where a score of -1 corresponds to the worst possible graph (the reverse of the true graph), a score of 1 represents a perfect match between the two graphs, and a score of 0 indicates a graph that has the same score as an empty or a fully connected graph.

\section{Evaluation and discussion of the results}
\label{results}
We begin the evaluation in subsection \ref{inference} by investigating inference and predictive validation, and in subsection \ref{graph} we evaluate the similarities between graphical structures learnt from the data and those obtained through domain knowledge. We then investigate the differences in the effect of intervention produced by some of these structures in subsection \ref{interventional analysis}, and perform sensitivity analysis in subsection \ref{sensitivity}. Finally, in subsection \ref{causal}, we undertake a qualitative causal evaluation of the graphical structures. 
Before we discuss the results, we present the model-averaging graph in Figure 10, which was produced as described in subsection \ref{model-average}. This average graph offers a visual representation of the combined insights from the 11 structure learning algorithms, and it is considered as part of the evaluation in addition to the 11 individual graphical structures learnt by each of the structure learning algorithms considered. Note that the average graph distinguishes itself by the absence of undirected edges, effectively addressing issues related to the directionality commonly associated with PDAGs and CPDAGs. This characteristic enhances the clarity of the graphical representation, contributing to a more refined and conclusive depiction of causal relationships within the model-averaging framework.

\begin{figure*}
\centering
    \includegraphics[width=\textwidth,height=6cm]{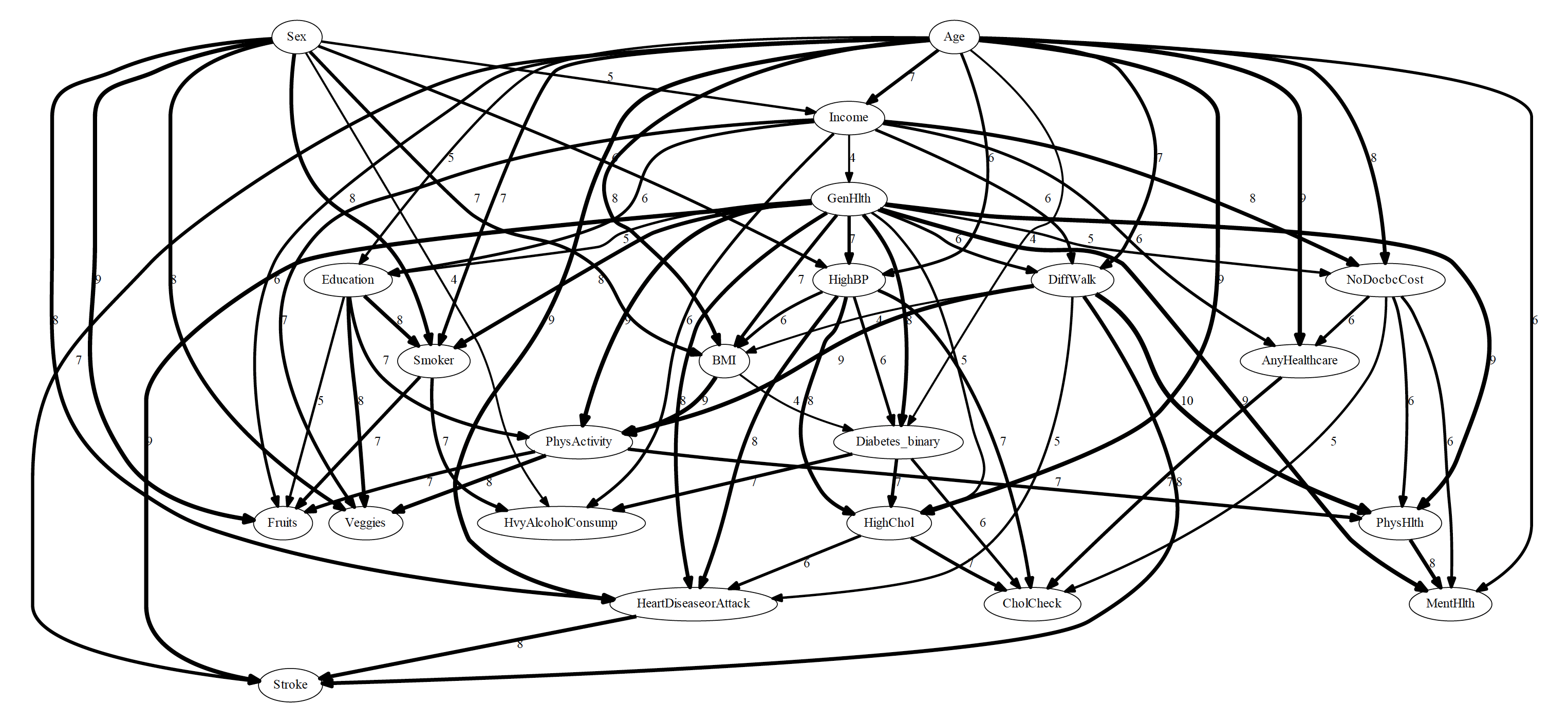}
    \caption{The model-averaging graph depicting the frequency of directed edges obtained from 11 structure learning algorithms. Edges appearing less than four times (i.e., appearing in less than a third of the graphs) across the 11 learnt outputs are excluded. The graph contains 75 directed edges across 22 nodes.}
  
\end{figure*}
\subsection{Inference-based evaluation and Predictive validation}
\label{inference}
The performance of the learnt CBNs is evaluated using two key metrics: the Bayesian Information Criterion (BIC) and Log-Likelihood (LL). BIC plays a crucial role in model selection by balancing model fit with complexity. It penalises models with increased complexity, thereby mitigating the risk of overfitting the data. In contrast, LL quantifies the degree to which the model reflects the observed data. More specifically, it measures the probability of observing the given data, assuming the model's underlying structure holds true. Higher LL values indicate a better fit between the model and the data.

\begin{figure*}[ht]
\centering
  \includegraphics[width=10cm]{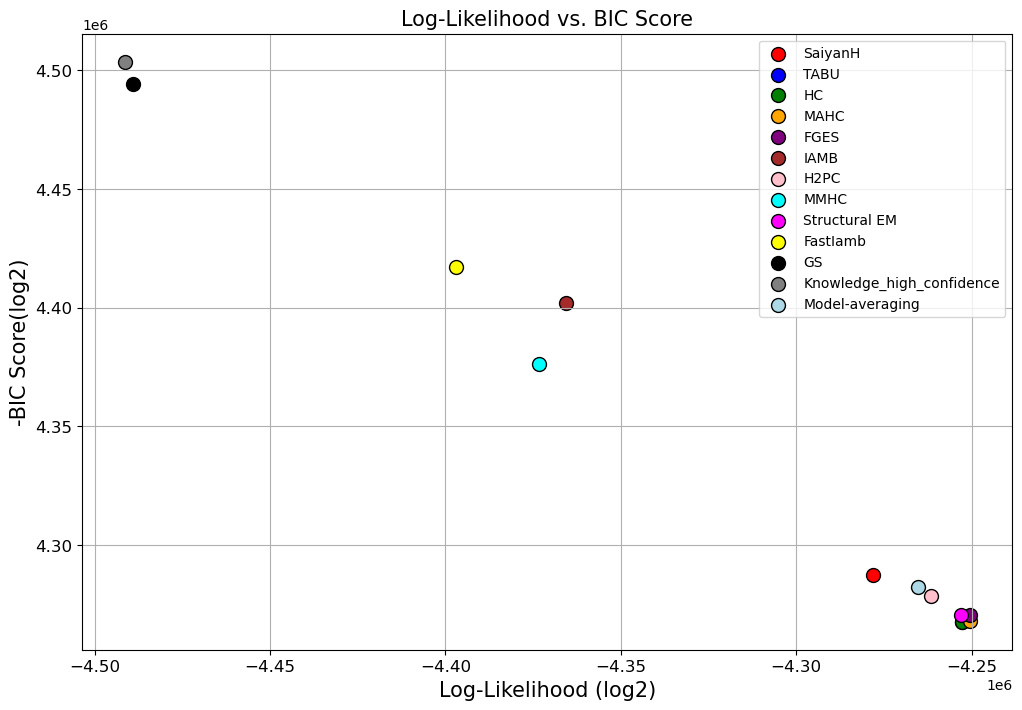}
  \caption{A two-dimensional scatter plot on the relationship between log-likelihood and the BIC score.}
   \label{fig:11}
\end{figure*}

Figure \ref{fig:11} illustrates the performance of various learning algorithms based on their LL and BIC scores. MAHC, FGES, HC, Structural EM and TABU emerge as top performers, achieving a desirable balance between model fit and complexity. H2PC demonstrates LL and BIC scores comparable to top performers. Notably, the model-averaging graph prioritises model balance, achieving a strong BIC score while retaining acceptable LL. IAMB, its variant FastIAMB, and MMHC generate moderate LL and BIC scores. Lastly, GS and the high confidence knowledge graph stand out as outliers due to their exceptionally low BIC and LL scores, relative to the scores produced by the other algorithms.GS, in particular, tends to produce sparse models. A low LL score typically indicates a reduced ability of the model to capture the underlying data complexity. Consequently, this often results in fewer edges, reflecting a simpler structure with potentially less explanatory power. Regarding the knowledge graph, its lower BIC score compared to others is unsurprising, as its primary purpose lies in representing existing knowledge rather than maximising model selection scores.

\begin{figure*}[ht]
\centering
  \includegraphics[width=10cm]{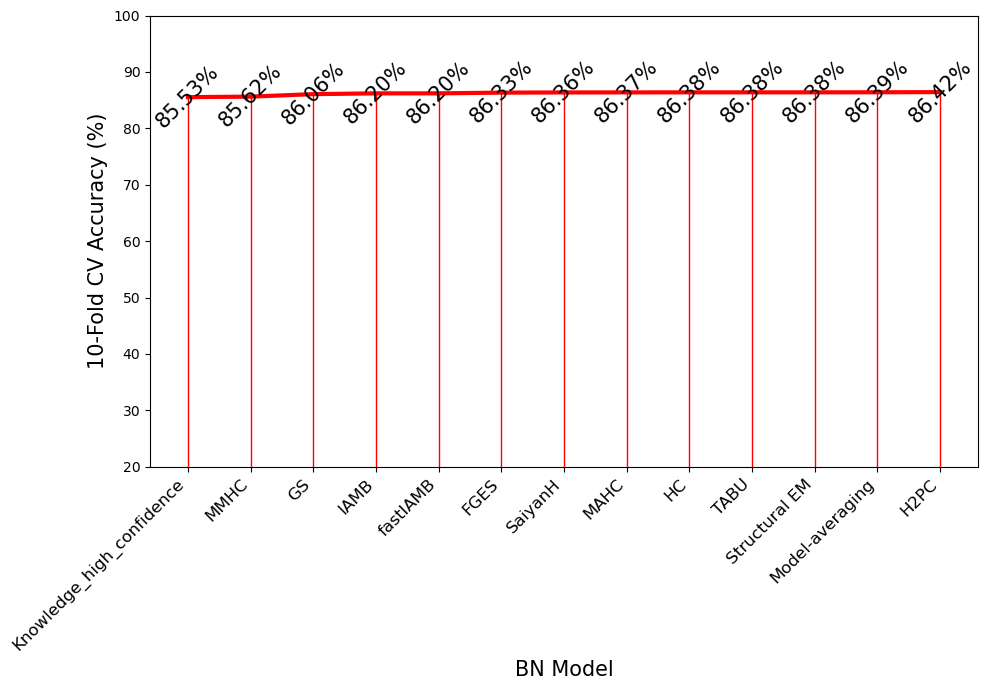}
  \caption{10-fold cross-validation classification accuracy achieved for predicting the target variable, Diabetes\_binary, across all models, ordered by worse to best performance.}
     \label{fig:12}
\end{figure*}

we also evaluated the performance of the learnt BN models using 10-fold cross-validation within the GeNIe BN software \cite{genie}. Figure 12 presents the classification accuracy achieved for predicting the target variable, Diabetes\_binary, across all models. It is important to note that predictive validation is generally not recommended for models that involve unsupervised learning, such as the causal structures discussed in this paper. This is because they do not optimise with respect to a specific target variable. Therefore, the results presented in Figure \ref{fig:12} should be interpreted with caution.

Our cross-validation analysis reveals that H2PC emerges as the frontrunner, achieving the highest accuracy in predicting unseen data. Closely following are Model-averaging and HC, demonstrating strong performance. However, it is observed that there is no meaningful difference in predictive accuracy across the various models, despite their fundamentally different causal structures. This finding aligns with previous research \cite{openproblems} indicating that predictive validation alone is insufficient for uncovering underlying causal distinctions. Instead, these differences can only be revealed through interventional analyses. On the other hand, these predictive results suggest the effectiveness of these models in generalising and making accurate predictions on data not used for training.

\subsection{Graphical structure}
\label{graph}

\
\begin{figure*}[ht]
\centering
  \includegraphics[width=12cm]{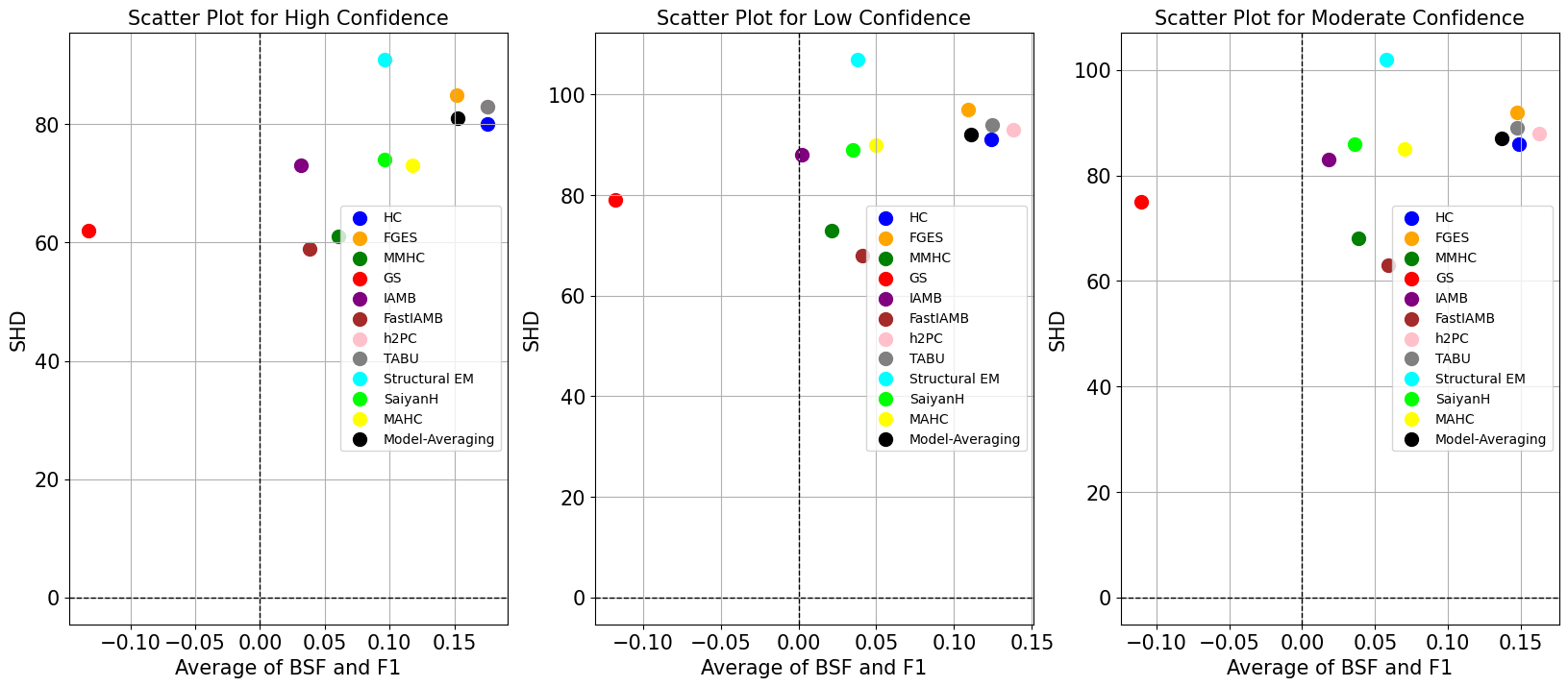}
  \caption{The SHD and the average of BSF+F1 scores produced by the 11 structure learning algorithms and the model-averaging graph, with reference to the three knowledge graphs reflecting the three different confidence levels.}
     \label{fig:13}
\end{figure*}

Figure \ref{fig:13} plots the SHD score relative to the average of F1 and BSF scores for each graph learnt from data (including the model-average graph) and with reference to the expert-produced knowledge graphs across the three confidence levels. Recall that higher SHD scores indicate lower similarity between graphs, whereas higher F1 and BSF scores indicate higher similarity.
\begin{figure*}[ht]
\begin{center}
    \includegraphics[width=12cm]{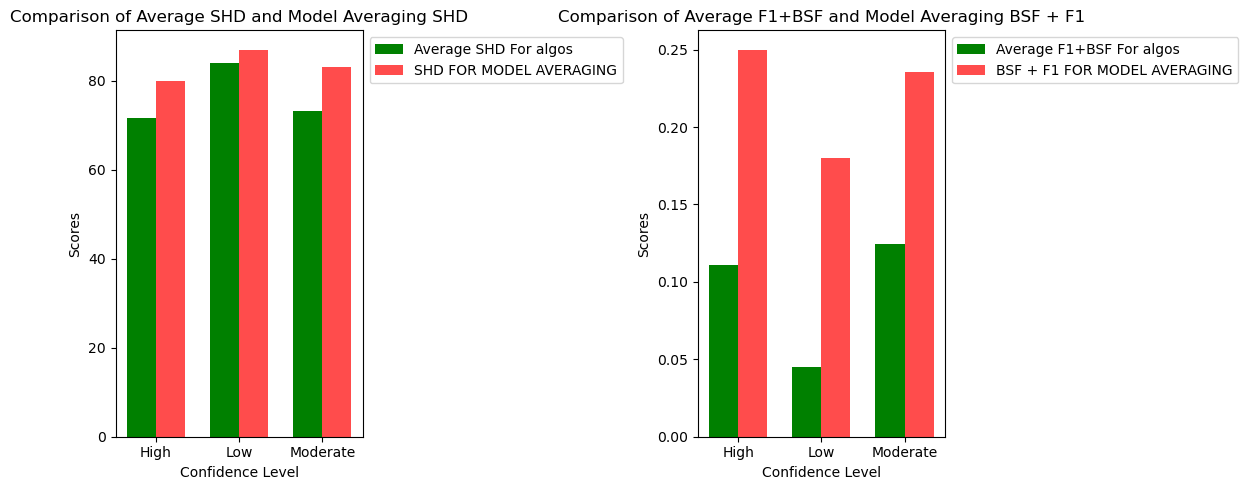}
    \caption{Comparison of average SHD and F1+BSF scores for different confidence levels with corresponding scores for model-average graph}
    \label{fig:14}
     \end{center}
\end{figure*}

We investigate the results further by comparing the average SHD and BSF+F1 scores across all algorithms for each confidence level to the corresponding SHD and BSF+F1 scores of the model-average graph. We present these results in Figure \ref{fig:14}. Overall, the results show that the independent algorithms produce an average SHD that is lower than that produced by the model-averaging graph, suggesting a closer alignment with the expert-produced knowledge graphs across the three confidence levels than the model-averaging graph. However, the model-averaging graph outperforms all 11 independently learnt graphical structures in terms of the F1 and BSF scores, indicating that it better captures the overall balance between precision and recall, as well as the difficulty of identifying both edge presence and absence. This disagreement between the SHD score and the F1 and BSF metrics is documented in \cite{DBLP}, and is attributed to the SHD score being biased towards sparser graphs. Specifically, the SHD score primarily reflects pure classification accuracy, whereas the F1 and BSF metrics encompass partly and fully balanced scores, respectively. This bias also explains why the SHD score undervalues the average graph compared to the F1 and BSF scores, which contains a higher number of edges compared to most of the independent graphs learnt by the 11 algorithms.

\subsection{Interventional analysis}
\label{interventional analysis}

We conduct intervention analyses targeting High Blood Pressure (HighBP), High Cholesterol (HighChol), Heart Disease or Heart Attack (HeartDiseaseorAttack), Body Mass Index (BMI), Education level (Education), and Physical Activity (PhysActivity) as the variables for intervention. Our aim is to assess the differences in potential influence of these interventions on diabetes (node called Diabetes\_binary) and investigate differences across graphical structures. 
\begin{figure*}[ht]
\centering
 \includegraphics[width=14cm]{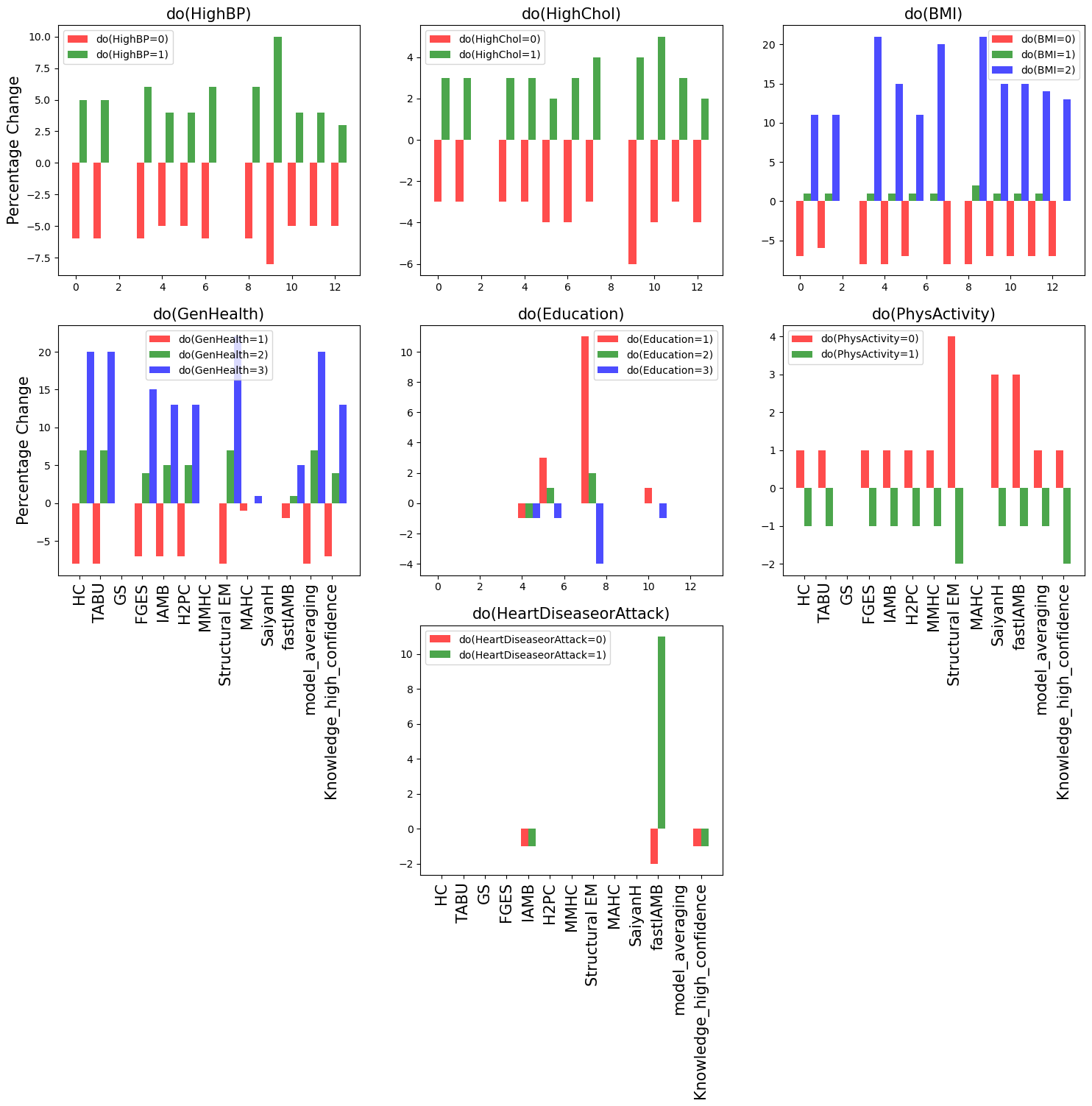}
    \caption{Percentage change of Diabetes\_binary given interventions on HighBP, HighChol, HeartDiseaseorAttack, BMI, Education, and PhysActivity.}
    \label{fig:15}
    
\end{figure*}

Figure \ref{fig:15} presents a comprehensive analysis of the percentage shift in diabetes resulting from interventions for the specified key variables. Overall, the results highlight important alterations in the likelihood of diabetes following interventions for High Blood Pressure (HighBP), High Cholesterol (HighChol), Body Mass Index (BMI), General Health (GenHealth), and Physical Activity (PhysActivity). Conversely, interventions targeting Education and HeartDiseaseorAttack exhibit minimal to negligible impact on the likelihood of diabetes across various graphs. 
The impact of interventions on diabetes, as observed across the majority of individual algorithms, consistently aligned with the patterns outlined in the knowledge graph, as illustrated in Figure \ref{fig:15}. Notably, the model-averaging graph demonstrated a tendency to converge towards the knowledge graph. Moreover, the impact of interventions appears to be partially linked to the density of graphs. This is explained by denser network facilitating increased propagation of information throughout the graph.

\begin{figure*}[ht]
\begin{center}

    \includegraphics[width=15cm]{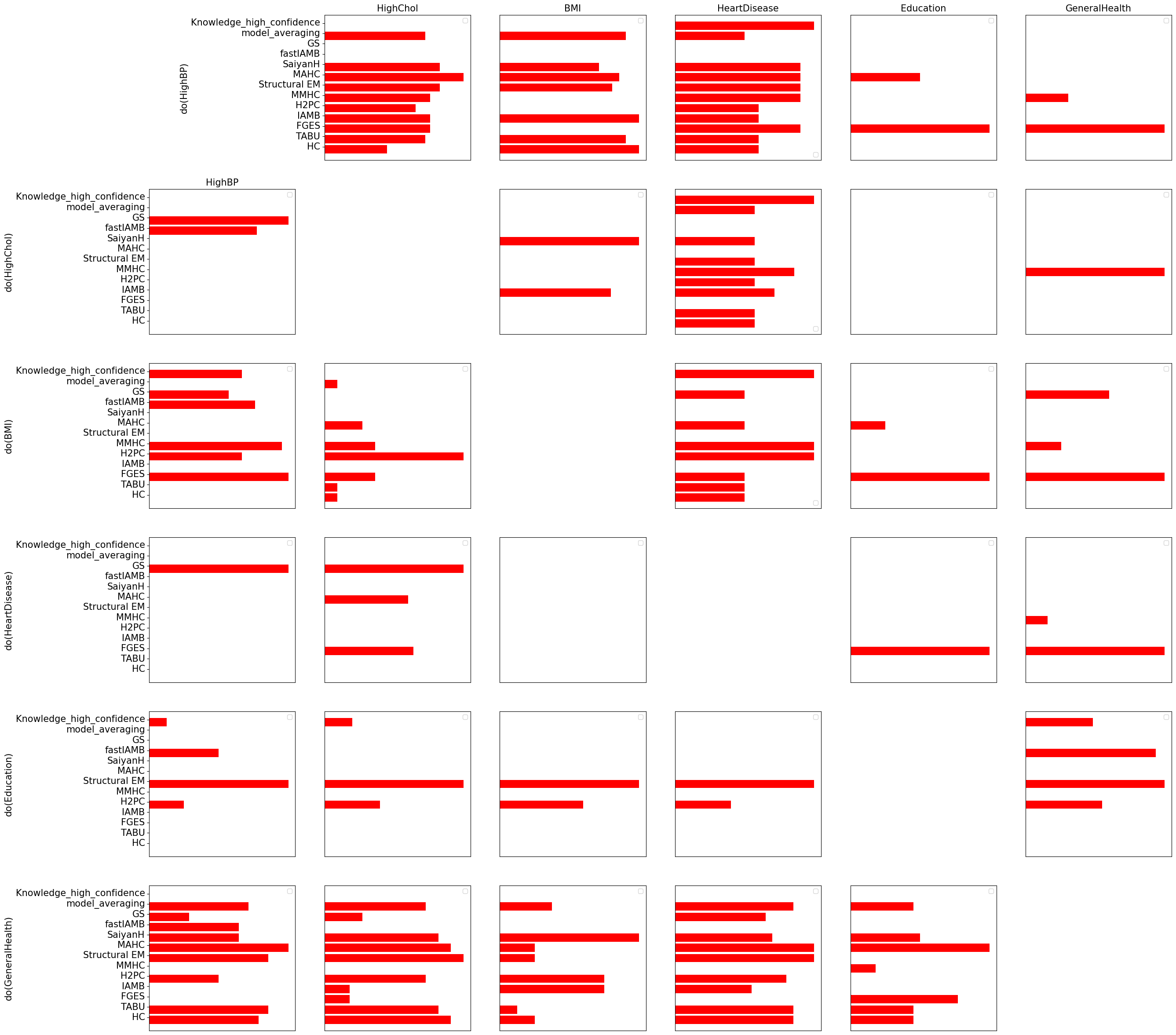}
    \caption{Illustration of the effects of specified hypothetical interventions on key variables. Results are presented through various algorithmically learnt graphs, the model-averaging graph, and the high-confidence knowledge graph. Absence of a result/red bar indicates no observed effect.}
   
    \label{fig:16}
     \end{center}
\end{figure*}

In addition to investigating the impact of interventions on diabetes, we extend our analysis to examine the effect on other variables. Figure \ref{fig:16} offers a comprehensive visual depiction of the effects of interventions on various factors across algorithmically learnt, model-averaging, and high-confidence knowledge graphs. Some key observations include:

\begin{enumerate}
\item \textbf{Intervening on HighBP \textit{do}(HighBP):} A notable impact of intervening on HighBP is observed through its influence on HighChol, BMI, and HeartDisease. This observed effect remains consistent across various algorithmically learnt graphs, the model-averaging graph, and the high-confidence knowledge graph. The relationship between HighChol and HighBP is complex, with indications of bidirectionality, which led to its absence in the high-confidence knowledge graph. However, given the consistent effects observed in intervening on HighBP, this suggests a potential starting point for further exploration. Another significant relationship emphasised is that between HighBP and HeartDisease.  This relationship has been well-documented, as evidenced by the Systolic Blood Pressure Intervention Trial (SPRINT)\cite{williamson2016intensive} which provided compelling evidence endorsing the efficacy of blood pressure reduction in preventing heart disease. 
\item \textbf{Intervening on HighChol \textit{do}(HighChol):} Intervening on HighChol has a strong impact on HeartDisease, while displaying minimal influence on other health factors. The relationship between HighChol and HeartDisease is well-documented in the literature. The findings align with existing clinical knowledge and guidelines related to the management of high cholesterol and its implications for the prevention and management of heart disease \cite{cdccholesterol}.
\item \textbf{Intervening on BMI \textit{do}(BMI):} This results in a significant effect on HeartDisease and HighBP. Research has consistently shown that an elevated BMI is associated with an increased risk of developing heart disease and experiencing adverse cardiovascular outcomes. Several studies and scientific statements from reputable sources, including the American Heart Association (AHA) \cite{powell2021obesity} and the British Heart Foundation (BHF) \cite{bhfobesity}, have highlighted the strong relation between BMI and HeartDisease.
\item \textbf{Intervening on HeartDisease \textit{do}(HeartDisease):} This appears to have a minimal impact on various factors, as evident from the figure.
\item \textbf{Intervening on Education \textit{do}(Education):} The impact of intervening on Education is notably insignificant across various health factors, as depicted in the figure. Existing research indicates that the influence of education on health is intricate and interwoven with various socioeconomic and behavioural factors \cite{education}. Therefore, the complex and multifaceted nature of this relationship justifies the notably insignificant impact of intervening on education across various health factors. 
\item \textbf{Intervening on GenHealth \textit{do}(GenHealth):} This exhibits a strong impact on HighBP, HighChol, BMI, HeartDisease, and Education as can be observed consistently across most of the learnt graphs. It highlights the pervasive influence of general health on overall well-being and its interconnectedness with various aspects of health and lifestyle.
\end{enumerate}

\subsection{Sensitivity Analysis}
\label{sensitivity}

Sensitivity analysis examines the responsiveness of a node to changes in its parent and ancestor nodes. High sensitivity implies that small adjustments to the Conditional Probability Table (CPT) parameters strongly impact posterior distributions, partly indicating a stronger dependency. Conversely, low sensitivity suggests that substantial changes to CPT parameters result in weak effects on posterior distributions, partly signifying a weaker dependency. As before, our focus is on the variable of interest Diabetes\_binary, to gain insights into its sensitivity characteristics across the different graphical structures constructed. We use the GeNIe BN software \cite{genie} to perform sensitivity analysis.
\begin{figure*}[ht]
\begin{center}

    \includegraphics[width=\textwidth]{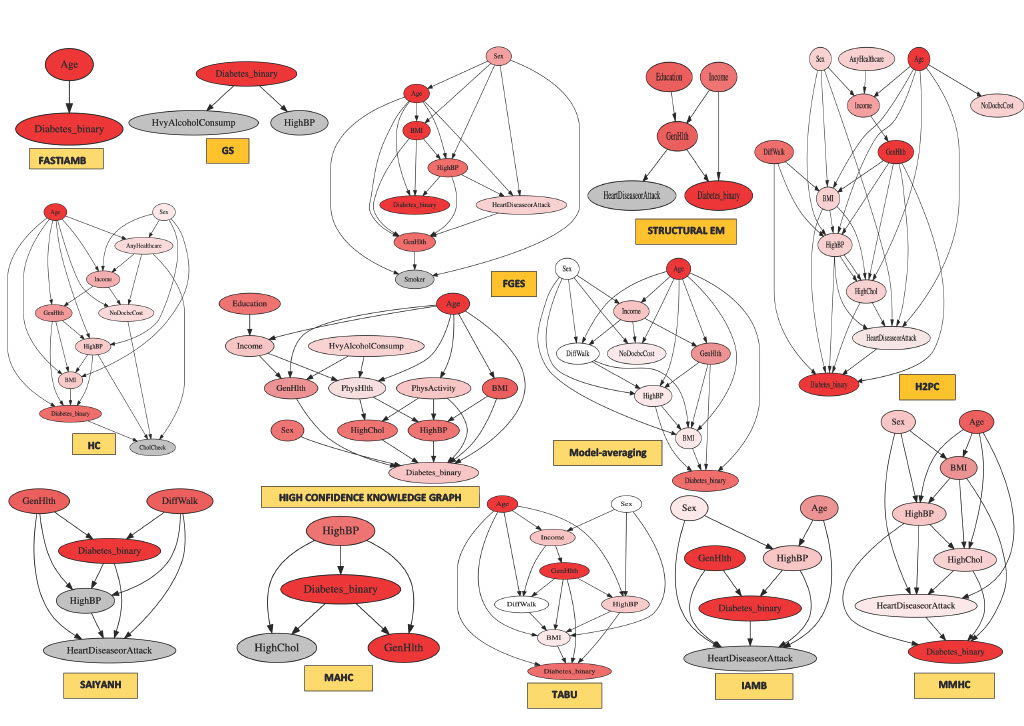}
    \captionsetup{position=bottom}
    \caption{Sensitivity analysis on the target node Diabetes\_binary, across the 11 graphs learnt by the algorithms, the model-average graph, and the high-confidence knowledge graph. A stronger red colour indicates higher sensitivity.}
    \label{fig:sensitivity}
    \end{center}
  \end{figure*}

Figure \ref{fig:sensitivity} illustrates the sensitivity analysis of the target node 
Diabetes\_binary considering the graphs derived from domain knowledge (high-confidence), the model-averaging approach, and each of the graphs learnt by the 11 structure learning algorithms. Only relevant fragments of these graphs are presented. Nodes not included in the set of parent and ancestor nodes of the target variable are shaded in grey, while the rest are highlighted in red where a darker red shade indicates stronger sensitivity.

The diverse outcomes derived from our sensitivity analysis provide insights into the multifaceted nature of the factors influencing Diabetes\_binary. GS indicates that Diabetes\_binary exhibits insensitivity to all other nodes. In contrast, algorithms such as H2PC, FGES, HC, TABU, model-averaging, and the high-confidence knowledge graph suggest that Diabetes\_binary is predominantly sensitive to Age. 
On the other hand, IAMB, StructuralEM, MAHC, and SaiyanH show that Diabetes\_binary displays heightened sensitivity to GenHlth. Moreover, H2PC and model-averaging graph indicate sensitivity to both Age and GenHlth, whereas FGES suggests sensitivity to Age and BMI. 
\\ Based on the results obtained, it can be deduced that diabetes is primarily sensitive to age, a factor that aligns with the existing literature, which emphasises the crucial role of age as a primary risk factor in the development of diabetes. It is widely acknowledged that the incidence of diabetes escalates with progressive age \cite{mordarska2017diabetes}.
GenHlth and BMI emerge as additional factors wherein the sensitivity of diabetes becomes apparent. This  aligns with existing research emphasising BMI's role as a risk factor for diabetes and its associated complications \cite{Luchsinger2023}. Increased BMI is associated with a higher likelihood of developing diabetes.

\subsection{Causal Relationship evaluation}
\label{causal}
We also perform qualitative assessment of the causal connections discovered by the structure learning algorithms. In this part of the evaluation, our attention is directed towards relationships that are clearly understood and are present in the high-confidence knowledge graph, as identified by the domain expert. This qualitative evaluation serves as a crucial step in validating the accuracy and relevance of the learnt causal structures, as well as investigating any strong disagreements between what the structure learning process suggests and what the domain expert indicates.

\begin{table*}[ht]
\small
\centering
\caption{Qualitative assessment of causal relationships.} 
\label{tab:relationships}
\begin{tabular}{llll}
\hline
\textbf{Parent} & \textbf{Child} & \textbf{Graphs by 11 algorithms} & \textbf{Model-averaging graph}\\
\hline

HvyAlcoholConsump  & PhysHlth & \cellcolor{red!50}No & \cellcolor{red!50}No \\
Sex  & Diabetes\_binary & \cellcolor{red!50}No & \cellcolor{red!50}No\\
Age & Income & \cellcolor{green!25} & \cellcolor{green!50}Yes\\
Age & GenHlth & \cellcolor{red!25} & \cellcolor{red!50}No\\
Age & Diabetes\_binary & \cellcolor{green!25} & \cellcolor{green!50}Yes\\
Age  & BMI &  \cellcolor{green!25} & \cellcolor{green!50}Yes\\
Age  & PhysActivity & \cellcolor{red!50}No & \cellcolor{red!50}No \\
Age  & DiffWalk & \cellcolor{green!25} & \cellcolor{green!50}Yes\\
Age  & PhysHlth & \cellcolor{red!50}No & \cellcolor{red!50}No \\
Income  & GenHlth & \cellcolor{green!25} & \cellcolor{green!50}Yes\\
Income & MentHlth   & \cellcolor{red!50}No & \cellcolor{red!50}No \\
Income & PhysHlth  & \cellcolor{red!50}No & \cellcolor{red!50}No \\
Income & AnyHealthcare & \cellcolor{green!25} & \cellcolor{green!50}Yes\\
Income & NoDocbcCost & \cellcolor{green!25} & \cellcolor{green!50}Yes\\
Income  & CholCheck & \cellcolor{red!50}No & \cellcolor{red!50}No \\
HvyAlcoholConsump &  MentHlth & \cellcolor{red!50}No & \cellcolor{red!50}No \\
HvyAlcoholConsump & GenHlth& \cellcolor{red!50}No & \cellcolor{red!50}No \\
AnyHealthcare & NoDocbcCost & \cellcolor{green!25} & \cellcolor{green!50}Yes(Reverse)\\
AnyHealthcare & CholCheck & \cellcolor{green!25} & \cellcolor{green!50}Yes\\
BMI & Diabetes\_binary& \cellcolor{green!25} & \cellcolor{green!50}Yes\\
BMI & HighBP & \cellcolor{green!50}Yes & \cellcolor{green!50}Yes\\
HighBP & HeartDiseaseorAttack & \cellcolor{green!25} & \cellcolor{green!50}Yes\\
HighBP & Stroke & \cellcolor{red!25} & \cellcolor{red!50}No\\
HighBP & Diabetes\_binary& \cellcolor{green!25} & \cellcolor{green!50}Yes\\
HighChol & HeartDiseaseorAttack & \cellcolor{green!25} & \cellcolor{green!50}Yes\\
HighChol & Diabetes\_binary & \cellcolor{green!25} & \cellcolor{green!50}Yes(Reverse)\\
GenHlth & Diabetes\_binary & \cellcolor{green!50}Yes & \cellcolor{green!50}Yes\\
GenHlth  & MentHlth & \cellcolor{green!25} & \cellcolor{green!50}Yes\\
Diabetes\_binary &  DiffWalk & \cellcolor{red!25} & \cellcolor{red!50}No\\
Education & Income & \cellcolor{green!50}Yes & \cellcolor{green!50}Yes(Reverse)\\
PhysActivity & Diabetes\_binary & \cellcolor{red!50}No & \cellcolor{red!50}No \\
PhysHlth  & MentHlth & \cellcolor{green!25} & \cellcolor{green!50}Yes\\
PhysHlth  & HighBP & \cellcolor{red!50}No & \cellcolor{red!50}No \\
PhysHlth  & HighChol & \cellcolor{red!50}No & \cellcolor{red!50}No \\
PhysActivity  & HighChol & \cellcolor{red!50}No & \cellcolor{red!50}No \\
PhysActivity  & HighBP & \cellcolor{red!25} & \cellcolor{red!50}No\\
HeartDiseaseorAttack & Stroke & \cellcolor{green!25} & \cellcolor{green!50}Yes\\
\hline
\end{tabular}
\end{table*}

In Table \ref{tab:relationships}, the varying  shades of red and green are used to highlight whether the graphs produced by the 11 algorithms and the model-averaging graph capture the knowledge-based relationships identified with high confidence by the domain expert. For the average graph, a red colour indicates disagreement and a green colour agreement with the specified knowledge-based edges, whereas different shades of red and greed are used to capture the level of agreement between the knowledge-based edges and the 11 graphs learnt by the different algorithms. Despite noticeable inconsistencies in learnt structures between algorithms, Table \ref{tab:relationships} demonstrates that structure learning algorithms have generally performed well in identifying many of the relationships provided in the high-confidence knowledge graph. 

The model-averaging graph aligns in 20 out of 37 edges with the high-confidence knowledge graph, and differs in the remaining 17 edges. This finding suggests that certain relationships are consistently predicted across different algorithms, while others show varying predictions. This highlights the importance of carefully selecting the appropriate algorithm for each specific study, as different algorithms may capture different aspects of the relationships between variables. In our investigation, we operate under the assumption of an equal contribution from each structure learning algorithm to compose the average graph. However, considering a weighted average that prioritises edges learnt by algorithms acknowledged for their heightened accuracy could be advantageous. Moreover, the coexistence of  agreement and disagreement in the model-averaging graph indicates that the relationships between the variables might be difficult to measure and involve complex interactions.

There are several areas of disagreement, particularly with regard to the relationships between sex and diabetes, age and physical activity, physical activity and diabetes, physical health and high blood pressure. While it is reasonable to assume that the lack of a direct relationship between sex and diabetes in the results may be attributed to the fact that sex influences complex biological processes within the human body, such as hormonal variations, genetic predispositions, and metabolic differences, which could potentially serve as the driving factors behind diabetes risk \cite{RUSSO20222297}, the dataset does not explicitly include these biological variables. As a result, the structure learning algorithms may fail to detect a direct relationship between sex and diabetes. The literature supports the notion that sex differences in metabolic regulation and diabetes susceptibility are influenced by both biological and psychosocial factors \cite{sexdiff} \cite{KautzkyWiller2016SexAG}.

The influence of age on physical activity is well-documented in existing research. As people age, they typically experience a reduction in physical activity, muscle mass, and muscle strength \cite{westerterp2018changes}. The Centers for Disease Control and Prevention \cite{cdcActive} emphasises the importance of physical activity for individuals with diabetes, and the American Diabetes Association's \cite{colberg} statement highlights that physical activity improves blood glucose control in type 2 diabetes and improves overall well-being. The absence of a direct causal link between the above variables could be explained by bidirectional influences, given the complex interactions within the context of these relationships. This means that we may need to consider multiple other factors in order to fully comprehend the underlying causal relationship between some of the variables considered in this study.
\section{Conclusion}
\label{conclusion}
This study investigated the discovery of causal graphical structures and the construction of CBNs from data on diabetes collated from BRFSS, which is a system of health-related telephone surveys that collect state data on risk behaviours, chronic health conditions, and use of preventative treatments amongst U.S. residents. This study contributes to the development and evaluation of CBNs for diabetes in different ways:
\begin{itemize}
\item[a)] Explores the distinctions amongst 11 structure learning algorithms in the context of discovering causal structures for diabetes.
\item[b)] Employs model-averaging of the outputs generated by diverse structure-learning algorithms to derive a unified and robust graphical structure, providing a more reliable representation of the causal relationships within the dataset.
\item[c)] Elicits causal relationships from expert knowledge and classifies knowledge-based edges into different levels of confidence as specified by a domain expert.
\item[d)] Performs a comparative analysis between the graphs obtained from structure learning algorithms and the knowledge graphs proposed by domain experts to identify similarities and disparities and enhance the understanding of inferred causal relationships in the context of diabetes.
\item[e)] Transforms the learnt structures into CBNs and assesses the potential effect of hypothetical interventions as well as the interactions between relevant risk factors.
\end{itemize}

To promote future research and collaboration, we make all the learnt graphs, the average graph, the three knowledge graphs, the CBN models of the average graph and the high-confidence knowledge graph, as well as the pre-processed dataset, publicly accessible through the Bayesys repository \cite{constantinoubayesys}.

The results highlight that the inferred causal structure is highly sensitive to the selection of the structure learning algorithm, which motivated our exploration of a model-averaging strategy to derive a representative graph from the diverse algorithmic outputs. Notably, our findings indicate that the model-averaged graph converged closer to the high-confidence knowledge graph, compared to the individual graphs learnt by the different algorithms.  Although model-averaging is found to effectively address variability between algorithms, it is crucial to acknowledge that the resultant average graph still exhibits notable distinctions compared to the knowledge graph, and this level of disagreement requires further investigation.  
 
In our exploration of the effects of hypothetical interventions on diabetes, we find that risk factors related to blood pressure, high cholesterol, BMI, general health, and physical activity, display a remarkable degree of similarity across the majority of the algorithms used in our study and show a parallel pattern of similarity in the high-confidence knowledge graph. Importantly, these observations align with the existing literature \cite{naha2021hypertension}, \cite{cholesterol}, \cite{klein2022obesity}  which highlights the role of these factors in diabetes. These experiments also emphasise the important influence of these variables on the likelihood of diabetes, reinforcing their significance in line with established scientific knowledge.

However, the results also highlight many inconsistencies between causal structures learnt from data and elicited from domain knowledge, and many inconsistencies extend between algorithms. For example, the sensitivity analysis conducted on diabetes across diverse graphical structures highlights the complexity of the factors influencing the development of
diabetes, and while certain algorithms suggest heightened sensitivity to variables,
such as Age, GenHlth and BMI, others indicate insensitivity. Therefore, while our findings mark a significant step in unravelling the complex set of relationships between risk factors and diabetes, it is imperative to recognise that further work is required to refine the accuracy of the inferred causal relationships.

The identification of causal pathways linking various variables to potential causes of diabetes through learnt graphs not only enables early detection but also plays a pivotal role in implementing effective preventive measures, significantly reducing the risk of long-term complications associated with diabetes. However, a key limitation and area for future research is that the traditional causal discovery methods for constructing CBNs neglect the temporal dimension, such as distribution shifts and time-varying causal structure, thereby limiting our capacity to accurately identify causal relationships between variables over time. This prevented the incorporation of suggestions from domain experts about evolving connections between diabetes and general health or diabetes and physical health into the knowledge graph, owing to concerns about introducing cyclicity. This limitation is particularly significant in healthcare data, where temporal information is essential for comprehending the dynamic nature of health conditions from longitudinal healthcare databases that capture individuals' health trajectories over time, which offer a valuable resource for discovering and modelling causal relationships. 

Several limitations related to the data warrant consideration in this study. Notably, the BRFSS data lacks a clear distinction between Type 1 Diabetes Mellitus (T1DM) and Type 2 Diabetes Mellitus (T2DM), prompting the exploration of potential insights derived from segregating these cohorts. Investigating whether the identified risk factors exhibit consistency across these subtypes could enhance our understanding of diabetes etiology. Additionally, the absence of explicit information on medication usage is a notable limitation. Although the survey inquires about medication for blood pressure control, the potential influence of specific medications on observed connections between certain nodes (e.g., high blood pressure and cholesterol) remains unexplored. The variable of medication, however, has not been incorporated into the present study, signaling a gap in our analysis that could be addressed in future research endeavors.

 \section*{CRediT Authorship Contribution Statement}
\textbf{Sheresh Zahoor:} Conceptualization, Methodology, Data
 curation, Visualization,  Formal analysis, Investigation,
 Resources, Software, Validation, Writing – original draft, Writing – review and editing.
\textbf{Anthony C. Constantinou:} Conceptualization, Methodology,
Formal analysis, Supervision, Validation, Writing – review and editing.
\textbf{Tim M Curtis:} Conceptualization, Supervision, Writing – review and editing. 
\textbf{Mohammed Hasanuzzaman:} Conceptualization, Supervision, Writing – review and editing.
 \section*{Declaration of Competing Interest}
 The authors declare that they have no known competing financial
interests or personal relationships that could have appeared to influence
 the work reported in this paper.
\section*{Acknowledgments}
This work has been jointly supported by Science Foundation Ireland [Grant number 18/CRT/6222] and the European Union’s Horizon 2020 research and innovation programme[Grant agreement number 101017385]. Additionally, for the purpose of Open Access, the author has applied a CC BY public copyright licence to any Author Accepted Manuscript version arising from this submission. The valuable support provided by BayesFusion, LLC in providing access to their software tool, GeNIe Modeler, is also acknowledged. 
\section*{Data availability}
Data is already made available online through a repository.

\bibliographystyle{elsarticle-num} 
\bibliography{primary}

\end{document}